\def\year{2022}\relax
\documentclass[letterpaper]{article} 
\usepackage{aaai22}  
\usepackage{times}  
\usepackage{helvet}  
\usepackage{courier}  
\usepackage[hyphens]{url}  
\usepackage{graphicx} 
\usepackage{natbib}  
\usepackage{caption} 
\DeclareCaptionStyle{ruled}{labelfont=normalfont,labelsep=colon,strut=off} 
\usepackage{algorithm}
\usepackage{algorithmic}

\usepackage{subfigure}
\usepackage{amsmath}
\usepackage{amssymb}
\usepackage{amsthm}
\usepackage{amsfonts}

\usepackage{tikz}

\usepackage[hidelinks]{hyperref}

\usepackage{bm}
\def\E{\mathbb{E}}

\def\dist{\mathrm{dist}}
\def\DTW{\mathrm{DTW}}
\def\GAK{\mathrm{GAK}}

\newtheorem{assumption}{Assumption}

\newtheorem{definition}{Definition}
\newtheorem{lemma}{Lemma}
\newtheorem{theorem}{Theorem}

\newtheorem{proposition}{Proposition}
\usepackage{newfloat}
\usepackage{listings}
\floatstyle{ruled}
\newfloat{listing}{tb}{lst}{}
\floatname{listing}{Listing}
\title{Training Robust Deep Models for Time-Series Domain: \\ Novel Algorithms and Theoretical Analysis}
\author{
Taha Belkhouja, Yan Yan, Janardhan Rao Doppa
}
\affiliations{
School of EECS, Washington State University\\
\{taha.belkhouja, yan.yan1, jana.doppa\}@wsu.edu

}

\usepackage{bibentry}

\begin{document}

\maketitle

\begin{abstract}
Despite the success of deep neural networks (DNNs) for real-world applications over time-series data such as mobile health, little is known about how to train robust DNNs for time-series domain due to its unique characteristics compared to images and text data.
In this paper, we fill this gap by proposing a novel algorithmic framework referred as {\em {\bf RO}bust Training for {\bf T}ime-{\bf S}eries (RO-TS)} to create robust deep models for time-series classification tasks. 
Specifically, we formulate a {\em min-max} optimization problem over the model parameters by explicitly reasoning about the robustness criteria in terms of additive perturbations to time-series inputs measured by the global alignment kernel (GAK) based distance. 
We also show the generality and advantages of our formulation using the summation structure over time-series alignments by relating both GAK and dynamic time warping (DTW).
This problem is an instance of a family of compositional min-max optimization problems, which are challenging and open with unclear theoretical guarantee. 
We propose a principled stochastic compositional alternating gradient descent ascent (SCAGDA) algorithm for this family of optimization problems. Unlike traditional methods for time-series that require approximate computation of distance measures, SCAGDA approximates the GAK based distance {\em on-the-fly} using a moving average approach. We theoretically analyze the convergence rate of SCAGDA and provide strong theoretical support for the estimation of GAK based distance.
Our experiments on real-world benchmarks demonstrate that RO-TS creates more robust deep models when compared to adversarial training using prior methods that rely on data augmentation or new definitions of loss functions. We also demonstrate the importance of GAK for time-series data over the Euclidean distance. The source code of RO-TS algorithms is available at \href{https://github.com/tahabelkhouja/Robust-Training-for-Time-Series}{https://github.com/tahabelkhouja/Robust-Training-for-Time-Series}
\end{abstract}
\section*{Introduction}

Predictive analytics over time-series data enables many important real-world applications including mobile health, smart grid management, smart home automation, and finance. 
In spite of the success of deep neural networks (DNNs) \cite{wang2017time}, very little is known about the robustness of DNNs for time-series data. 
Recent work on adversarial examples for image \cite{kolter2018materials} and text data \cite{wang2019deep} exposed the brittleness of DNNs and motivated methods to improve their robustness. 
Therefore, training robust deep models is a necessary requirement when we deploy DNNs over time-series data in critical and high-stakes applications (e.g., mobile health \cite{belkhouja2020analyzing}).
Real-world deployment of deep models for time-series data pose several robustness challenges. 
First, labeled training set for time-series domain tend to be smaller when compared to image and text domains. 
Consequently, the learned DNNs may not perform well on unseen data drawn from the same distribution. 
Second, disturbances due to noisy sensor observations or different sampling frequencies can potentially result in poor performance of DNN models. 
Third, adversarial attacks for malicious purposes to break DNN models can pose security threats. 

Prior work on robustness for image data considers training DNNs to be robust to adversarial attacks in a $L_p$ ball and input perturbations.  Algorithms to improve robustness of DNNs fall into two broad categories. First, adversarial training using data augmentation (e.g., adversarial examples or input perturbations). Second, optimizing an explicit loss function for robustness criterion (e.g., similar input images should produce similar DNN outputs). Almost all prior methods are designed for images and are based on the $L_p$ norm distance. Since time-series data has unique characteristics (e.g., sparse peaks, fast oscillations), $L_p$ distance can rarely capture the true similarity between time-series pairs and prior methods are likely to fail on time-series data, as demonstrated by results in Fig. \ref{fig:advtrn}. The main question of this paper is: {\em what are good methodologies to train robust DNNs for time-series domain by handling its unique challenges?}

To answer this question, we propose a novel and principled framework referred as {\em {\bf RO}bust Training for {\bf T}ime-{\bf S}eries (RO-TS)} to create robust DNNs for time-series data. 
We employ additive noise variables to simulate perturbations within a small neighborhood around each training example. We incorporate these additive noise variables to formulate a {\em min-max} optimization problem to reason about the robustness criteria in terms of disturbances to time-series inputs by minimizing the worst-case risk. 
To capture the special characteristics of time-series signals, we employ the global alighnment kernel (GAK) based distance \cite{cuturi2007kernel} to define neighborhood regions for training examples. We show the generality and advantage of our formulation using the summation structure over time-series alignments by relating both GAK and dynamic time warping (DTW) \cite{berndt1994using}.

Unfortunately, the min-max optimization problem with GAK based distance is challenging and may fall in the family of compositional min-max problems due to lack of theoretical guarantees.
To efficiently solve this general family of optimization problems, we develop a principled {\em stochastic compositional alternating gradient descent ascent (SCAGDA)} algorithm by carefully leveraging the underlying structure of this problem. Another key computational challenge is that time-series distance measures including the GAK based distance involve going through all possible alignments between pairs of time-series inputs, which is expensive, e.g., $O(T^2)$ for GAK where $T$ is the length of time-series signal. As a consequence, the computational cost grows significantly for iterative optimization algorithms where we need to repeatedly compute the distance between time-series signals. 
SCAGDA randomly samples a constant number of alignments at each iteration to approximate the GAK based distance {\em on-the-fly} using a moving average approach and reduces the computational-complexity to $O(T)$.

Our theoretical analysis shows that SCAGDA achieves an $\epsilon$-primal gap in $O(1/\epsilon^2)$ iterations for a family of nonconvex-nonconcave compositional min-max optimization problems. {\em To the best of our knowledge, this is the first convergence rate established for nonconvex-nonconcave min-max problems with a compositional structure}. We also prove that SCAGDA approximates GAK based distance during the execution of algorithm: the approximation error converges as the primal gap converges. This result provides strong theoretical support for our algorithm design that leverages the structure of the RO-TS formulation. Our experiments on real-world datasets show that DNNs learned using RO-TS are more robust than prior methods; and GAK based distance has a more appropriate bias for time-series than $L_2$. 

\vspace{1.0ex}

\noindent {\bf Contributions.} The key contribution of this paper is the development and evaluation of the RO-TS algorithmic framework to train robust deep models for time-series domain.

\begin{itemize}
\setlength\itemsep{0em}

    \item Novel SCAGDA algorithm for a family of nonconvex-nonconcave compositional min-max problems that covers RO-TS as a special case with GAK based distance. SCAGDA approximates the GAK distance by randomly sampling a constant number of time-series alignments. 
    
    \item Theoretical analysis of SCAGDA shows that the iteration complexity to achieve $\epsilon$-primal gap and the $\epsilon$-approximation error of GAK distance is $O(1/\epsilon^2)$.
          
    \item Comprehensive experimental evaluation of RO-TS on diverse real-world benchmark datasets and comparison with state-of-the-art baselines. The source code of RO-TS algorithms is available at \href{https://github.com/tahabelkhouja/Robust-Training-for-Time-Series}{https://github.com/tahabelkhouja/Robust-Training-for-Time-Series}
\end{itemize}

\section*{Problem Setup and Formulation}

We consider the problem of learning {\em robust} DNN classifiers over time-series data. We are given a training set of $n$ input-output pairs $\left\{\left(x_i,y_i\right)\right\}_{i=1}^{n}$. 
Each input $x_i \in \mathcal X$ is a time-series signal, where $\mathcal X \subseteq \mathbb{R}^{C \times T}$ with $C$ denoting the number of channels and $T$ being the window-size of the signal; 
and $y_i \in \mathcal Y$ is the associated ground-truth label, where $\mathcal Y \in \{ 1,\cdots, \mathcal C\} $ is a set of $\mathcal C$ discrete class labels. 
For example, in a health monitoring application using physiological sensors for patients diagnosed with cardiac arrhythmia, we use the measurements from wearable devices to predict the likelihood of a cardiac failure. 
Traditional empirical risk minimization learns a DNN classifier $f: \mathcal X \times \Theta \rightarrow \mathcal Y$ 
with weights $w \in \Theta$ that maps time-series inputs to classification labels for a hypothesis space $\Theta$ and a loss function $\ell$:
\begin{align*}
\min_{w \in \Theta} \frac{1}{n} \sum_{i=1}^n \ell(f(x_i, w), y_i) .
\end{align*}

\vspace{0.5ex}

\noindent {\bf Training for robustness.} We would like the learned classifier $f(x, w)$ to be robust to disturbances in time-series inputs due to noisy observations or adversarial attacks. For example, a failure in the prediction task for the above health monitoring application due to such disturbances can cause injury to the patient without the system notifying the needed assistance. Therefore, we want the trained DNN classifier to be invariant to such disturbances. Mathematically, for an appropriate distance function $d(x, x')$ over time-series inputs $x$ and $x'$, we want the classifier $f$ to predict the same classification label as $x$ for all inputs $x'$ such that $d(x,x') < \varepsilon$, where $\varepsilon$ stands for the bound on allowed disturbance to input $x$. 
This goal can be achieved by reasoning about the worst-case empirical risk over possible perturbations $a_i \in \mathbb{R}^{C \times T}$ of $x_i$ such that $d(x_i, x_i+a_i) \leq \varepsilon$. 
The resulting {\em min-max} optimization problem is given below. 
\begin{align}\label{eq:robust_to_max_ball_perturbation_constraint}
\min_{w \in \Theta} ~&~ \frac{1}{n} \sum_{i=1}^n \max_{a_i} \ell(f(x_i + a_i, w), y_i )
\nonumber\\
\text{s.t.} ~&~ d(x_i, x_i + a_i) \leq \varepsilon
\end{align}

In practice, instead of solving the above hard constrained problem, one can solve an equivalent soft constrained problem using regularization as follows
\begin{align}\label{eq:regularized_minmax}
\min_{w \in \Theta} \max_{a_i}  \frac{1}{n} \sum_{i=1}^n \ell(f(x_i+a_i, w), y_i) - \lambda d(x_i, x_i+a_i)
\end{align}

There is a natural interpretation of this optimization problem. The {\em inner maximization} problem serves the role of an attacker whose goal is to find adversarial examples that achieves the largest loss. The {\em outer minimization} problem serves the role of a defender whose goal is to find the parameters of the deep model $w$ by minimizing the adversarial loss from the inner attack problem. 
This formulation is applicable to all types of data by selecting an appropriate distance function $d$. 
For example, $L_p$-norm distance is usually used in the image domain \cite{kolter2018materials}.

Typical stochastic approaches to solving the above adversarial training problem include alternating stochastic optimization \cite{yang2020global} and stochastic gradient descent ascent (GDA) \cite{lin2020near,yan2020optimal}. 
The alternating method first fixes $w$ and solves the inner maximization approximately to get each $a_i$ (e.g., using stochastic gradient descent). Next, $a_i$ is fixed and the outer minimization is solved over $w$. 
These two steps are performed alternatively until convergence. 
The GDA method computes the gradient of $w$ and $a_i$ simultaneously at each iteration, and then use these gradients to update $w$ and $a_i$.
Both methods require the ability to compute the unbiased estimation of the gradients w.r.t. $w$ and $a_i$.
When $d$ is decomposable, e.g., $L_p$-norm, then its stochastic gradient can be easily computed.
However, in time series domain, commonly used distance measures may not be decomposable, so its stochastic gradients are not accessible. 
Consequently, one has to calculate the exact gradient of $d(x_i, x_i+a_i)$ w.r.t, $a_i$. We will investigate this key challenge in Section \ref{subsection:SCAGDA}.

We summarize in Table \ref{tab:notation} the main mathematical notations used in this paper.
\begin{table}[t]
\centering
\begin{tabular}{|c|l|} 
\hline
\textbf{Variable}  & \textbf{Definition} \\ \hline
$\mathcal{X}$ & Input space of time-series data\\ \hline
$\mathcal{Y}$ & Discrete set of class labels \\ \hline
$f$ & Deep neural network classifier \\ \hline
$w \in \Theta$ & Weights of classifier $f$ \\ \hline
$l(\cdot,\cdot,\cdot)$ & Surrogate loss function \\ \hline
$k_{\GAK} (\cdot,\cdot)$ & Global alignment kernel \\ \hline
$\pi$ & Alignment between two time-series  \\& defined as a pair $(\pi_1, \pi_2)$\\ \hline
$\mathcal{A}$ & Set of all possible alignments \\ \hline
$d_{\pi}(\cdot,\cdot)$ & Distance function according  \\& to an alignment $\pi$\\ \hline
\end{tabular}
\caption{Mathematical notations used in this paper.}
\label{tab:notation}
\end{table}

\section*{RO-TS Algorithmic Framework}

In this section, we describe the technical details of our proposed RO-TS framework to train robust DNN classifiers for time-series domain. First, we instantiate the min-max formulation with GAK based distance as it appropriately captures the similarity between time-series signals. Second, we provide an efficient algorithm to solve the GAK-based formulation to learn parameters of DNN classifers.

\subsection*{Distance Measure for Time-Series}

Unlike images and text, time-series data exhibits unique characteristics such as sparse peaks, fast oscillations, and frequency/time shifting which are important for pattern matching and data classification. 
Hence, measures such as Euclidean distance that do not account for these characteristics usually fail in recognizing the similarities between time-series signals. 
To address this challenge, elastic measures have been introduced for pattern-matching tasks for time-series domain \cite{cuturi2007kernel}, where one time-step of a signal can be associated with many time-steps of another signal to compute the similarity measure. 

\vspace{0.5ex}

\noindent {\bf Time-series alignment.} Given two time series $x$=$(x_1,\cdots, x_{T_1})$ and $x'$=$(x'_1,\cdots, x'_{T_2})$ for $T_1, T_2 \in \mathbb N_+$, the alignment $\pi$ = $(\pi_1, \pi_2)$ is defined as a pair of increasing integral vectors of length $r \leq T_1 + T_2 - 1$ such that $1$ = $\pi_1(1) \leq \cdots \leq \pi_1(r)$ = $T_1$ and $1$ = $\pi_2(1) \leq \cdots \leq \pi_2(r)$ = $T_2$ with unitary increments and without simultaneous repetitions, which presents the coordinates of $x$ and $x'$. This alignment defines the one-to-many alignment between $x$ and $x'$ to measure their similarity. 
Using a candidate alignment $\pi$, we can compute their similarity as follows:
\begin{align}
\label{eq:dpi}
d_\pi(x, x') = \sum_{i=1}^{|\pi|} \dist( x_{\pi_1(i)}, x'_{\pi_2(i)} )
\end{align}
where $|\pi|$=$r$ denotes the length of alignment and $\dist(\cdot, \cdot)$ in the above equation is the Minkowski distance:
\begin{equation}
\dist( x_{\pi_1(i)}, x'_{\pi_2(i)} ) = \| x_{\pi_1(i)} - x'_{\pi_2(i)} \|_p , ~p\in\{1,\cdots,\infty\} \nonumber
\end{equation}

\vspace{0.5ex}

\noindent {\bf Global alignment kernel (GAK) based distance.} The concept of alignment allows us to take into consideration the intrinsic properties of time-series signals, such as frequency shifts, to compute their similarity. 
There are some well-known approaches to define distance metrics using time-series alignment. 
For example, dynamic time warping (DTW) \cite{berndt1994using} selects the alignment with the minimum distance:
\begin{align*}
D_{\DTW}(x, x') = \min_{\pi \in \mathcal A} d_\pi(x, x'),
\end{align*}
where $\mathcal A$ denotes the set of all possible alignments.

While DTW only takes into account one candidate alignment, global alignment kernel (GAK) \cite{cuturi2007kernel} takes all possible alignments into consideration:
\begin{align}\label{eq:GAK}
k_{\GAK} (x, x') 
= 
\sum_{ \pi \in \mathcal A } \exp\Big( - \frac{ d_\pi( x, x' ) }{ \nu } \Big)
\end{align}
where $\nu$ is a hyper-parameter and $d_\pi(\cdot, \cdot)$ is defined in Equation (\ref{eq:dpi}). 
In practice, to handle the diagonally dominance issue \cite{cuturi2007kernel,wu2018random,cuturi2011fast}, $D_{\GAK} := -\nu \log(k_{\GAK})$ is typically used as a distance measure for a pair of time-series signals.
GAK enjoys several advantages over DTW \cite{cuturi2011fast}:
(i) differentiable, (ii) positive definite, (iii) coherent measure over all possible alignments.
Therefore, $k_{\GAK}$ (or $D_{\GAK}$) is a better fit to train robust DNNs for the time-series domain.

On the other hand, GAK can also be a more {\it general} measure than DTW due to its summation structure, as $\lim_{\nu \rightarrow 0} D_{\GAK}(x, x') = D_{\DTW}(x, x')$, i.e., arbitrarily close to DTW by changing $\nu$.
The following proposition shows the tight approximation of the soft minimum of GAK to the hard minimum of DTW 
(proof and details in Appendix).

\begin{proposition}\label{prop:GAK_converge_to_DTW}
For a time-series pair $(x, x')$, we have:
\begin{align*}
0
\leq
D_{DTW}(x, x') - D_{\GAK}(x, x')
\leq 
\nu \log( | \mathcal A| )  .
\end{align*}
\end{proposition}
\noindent
As shown, $D_{\GAK}$ converges to $D_{\DTW}$ in $\nu \log(|\mathcal A|)$ as $\nu$ decreases. 
Due to the above advantages and the approximation ability of $D_{\GAK}$ to $D_{\DTW}$, we consider the more {\it general} $k_{\GAK}$ and $D_{\GAK}$ in our RO-TS method.

\subsection*{SCAGDA Optimization Algorithm}
\label{subsection:SCAGDA}

By plugging $k_{\GAK}$ from Equation (\ref{eq:GAK}) to replace $d$ into the min-max formulation in Equation (\ref{eq:regularized_minmax}), we reach the following objective function of our RO-TS framework:
\begin{align}\label{eq:ROTS}
\min_{w \in \Theta} \max_{a_i } &
\frac{1}{n} \sum_{i=1}^n \ell(f(x_i + a_i, w), y_i ) \nonumber\\
&
+ \lambda \overbrace{ \log\big( k_{\GAK} (x_i, x_i+a_i) \big) }^{=d(x_i, x_i+a_i)}
\end{align}
\noindent 
where $\nu$ outside $\log$ in $D_{\GAK}$ can be merged into $\lambda$.
The above problem is decomposable over individual training examples (i.e., index $i$), so we can compute stochastic gradients by randomly sampling a batch of data and employ stochastic gradient descent ascent (SGDA) \cite{lin2020near,yan2020optimal}, a family of stochastic algorithms for solving min-max problems. 

\vspace{0.5ex}

\noindent
{\bf Key challenge.} The second term  $\log(k_{\GAK}(x_i, x_i + a_i))$ has a {\it compositional} structure due to the outer $\log$ function.
By chain rule, its gradient w.r.t. the dual variable $a_i$ is
\begin{align*}
\nabla_{a_i} \log(k_{\GAK}(x_i, x_i + a_i))
=
\frac{ \nabla_{a_i} k_{\GAK}(x_i, x_i + a_i) }{ k_{\GAK}(x_i, x_i + a_i) }
\end{align*}
\noindent
where one has to go through all possible alignments to compute $k_{\GAK}$ and $\nabla_{a_i} k_{\GAK}$ (see Equation (\ref{eq:GAK})) and there is no unbiased estimation (i.e., stochastic gradients) for it.

Consequently, at each iteration, SGDA has to compute the exact value of $k_{\GAK}(x_i, x_i+a_i)$ and $\nabla_{a_i} k_{\GAK}(x_i, x_i + a_i)$ according to the chain rule, which leads to an additional time-complexity of $O(CT^2)$ per SGDA iteration, where $C$ and $T$ denote the number of channels and window-size respectively. 
This computational bottleneck will lead to extremely slow training algorithm when $C$ and/or $T$ is large, which is the case in many real-world applications.

One candidate approach to alleviate the computational challenge due to $\log(k_{\GAK})$ part of the objective is to make use of the inner summation structure of $k_{\GAK}$. 
Since $k_{\GAK}$ involves a summation over all alignments, as shown in (\ref{eq:GAK}), we can use only a {\it subset} of alignments for estimating the full summation. 
This procedure will give an unbiased estimation of $k_{\GAK}$, but the outer logarithmic function makes it a {\it biased} estimation for $\nabla_{a_i} \log(k_{\GAK}(x_i, x_i + a_i))$. 
However, such biased estimation violates the assumption in SGDA studies, so their theoretical analysis cannot hold. 

There is another line of research investigating stochastic compositional gradient methods for  minimization problems with compositional structure \cite{wang2017stochastic,chen2020solving}. 
However, {\em min-max} optimization with {\it compositional} structure, including our case shown in Equation (\ref{eq:ROTS}), is not studied yet. 
It is unclear whether these techniques and analysis hold for min-max problems.

\vspace{1.0ex}

\begin{algorithm}[t]
\caption{SCAGDA (Stochastic Compositional Alternating Gradient Descent Ascent)}
\label{algorithm:SCAGDA}
\begin{algorithmic}[1]
\STATE Initialize $w_0, a_i^0$ for $i=1,\cdots,n$ and $\omega_i^0 = 0$ for $i=1, \cdots, n$, step sizes $\{ \eta_k \}_{k=1}^K$ and $\{ \gamma_k \}_{k=1}^K$.

\FOR{$k = 0,\cdots, K-1$}

  \STATE Randomly sample an index $i_1$ to compute stochastic gradient $\nabla_w f_{i_1}(w_k, a_{i_1}^k)$
  \label{alg:line:primal_gradient}

  \STATE Set: $w_{k+1} = w_k - \eta_k \nabla_w f_i(w_k, a_{i_1}^k)$

  \STATE Randomly sample an index $i_2$ to compute stochastic gradient $\nabla_a f_{i_2}(w_{k+1}, a_{i_2}^k)$
  \label{alg:line:dual_gradient}

  \STATE Randomly sample two independent indices $j_1, j_2$ of $h_{i_2}$ to compute $h_{i_2, j_1}(a_{i_2}^k)$ and $\nabla h_{i_2, j_2}(a_{i_2}^k)$
  \label{alg:line:random_compositional_elements}

  \STATE Set:
  \begin{equation*}
      \omega_i^{k+1} = 
      \left\{
      \begin{array}{ll}
        \omega_i^k & \text{ for } i \neq i_2  \\
        (1-\beta) \omega^k_{i_2} + \beta h_{i_2, j_1}(a_{i_2}^k)  & \text{ for } i = i_2 
      \end{array}
      \right.
  \end{equation*}

  \label{alg:line:mv_compositional}

  \STATE Set: $a_{i_2}^{k+1} = a_{i_2}^k + \gamma_k ( \nabla_a f_{i_2}(w_{k+1}, a_{i_2}^k) - \nabla h_{j2}(a_{i_2}^k)^\top \nabla g(\omega^{k+1}_{i_2}) )$

\ENDFOR

\STATE {\bf return} final solution $w_K$
\end{algorithmic}
\end{algorithm}

\noindent {\bf SCAGDA algorithm.} We propose a novel {\em stochastic compositional alternating gradient descent ascent (SCAGDA)} algorithm to solve a family of nonconvex-nonconcave min-max compositional problems, which include RO-TS (Equation (\ref{eq:ROTS})) as a special case. We summarize SCAGDA in Algorithm \ref{algorithm:SCAGDA}. 
Specifically, we consider solving the following family of problems:
\vspace*{-0.2in}
\begin{align}\label{eq:min_max_general_form}
\min_w \max_{a_i} \frac{1}{n} \sum_{i=1}^n \phi_i(w, a_i)
\end{align}
where $\phi_i(w, a_i) := f_i(w, a_i) - g( \frac{1}{m} \sum_{j=1}^m h_{i,j} (a_i) )$. 

\vspace{1.0ex}

{\bf Mapping Problem (\ref{eq:min_max_general_form}) to RO-TS (\ref{eq:ROTS}).}
As mentioned above, RO-TS for time-series in Equation (\ref{eq:ROTS}) is a special case of Problem (\ref{eq:min_max_general_form}) as shown below.
The variables $\phi_i(w, a_i), f_i, g, h_{i,j}$ in Problem (\ref{eq:min_max_general_form}) can be instantiated by the following mappings:
\begin{itemize}
\item 
$f_i$ in $\phi_i$ of (\ref{eq:min_max_general_form}) 
$\Rightarrow$ 
the loss $\ell$ on the $i$-th data in (\ref{eq:ROTS})
\item 
$-g(\cdot)$ in $\phi_i$ of (\ref{eq:min_max_general_form}) $\Rightarrow$ $\lambda \log(\cdot)$ in (\ref{eq:ROTS})
\item 
$\frac{1}{m} \sum_{j=1}^m h_{i,j} (a_i)$ in $\phi_i$ of (\ref{eq:min_max_general_form}) $\Rightarrow$ $k_{\GAK}(x_i, x_i+a_i) = \sum_{\pi \in \mathcal A} \exp( - d_\pi (x_i, x_i + a_i) / \nu )$ in (\ref{eq:ROTS}), where $m$ and $j$ corresponds to the total number of alignment paths $|\mathcal A|$ and the index of alignment path, respectively.
Note that the summation form of $k_{\GAK}$ can be easily converted to an average form due to $\log(x) = \log(x/m) + \log(m)$.
\end{itemize}

\vspace{1.0ex}

{\bf Algorithmic analysis of SCAGDA.}
To introduce and analyze Algorithm \ref{algorithm:SCAGDA} for solving Problem (\ref{eq:min_max_general_form}), we first introduce some notations.
Denote $P(w) := \max_{a_i} \frac{1}{n} \sum_{i=1}^n \phi_i(w, a_i)$ as the {\it primal function} of the above min-max optimization problem, where we are interested in analyzing the convergence of the {\it primal gap} after the $K$-th iteration: $$P(w_K) - \min_w P(w).$$
Let $a := (a_1, a_2, ..., a_n) \in \mathbb R^{C \times T \times n}$ be the concatenation of $a_i$ for $i=1,\cdots, n$. We also use the following notations to improve the technical exposition and ease of readability. 
\begin{align*}
\phi(w,a) := & \frac{1}{n} \sum_{i=1}^n \phi_i(w, a_i),
\\
h_i(a_i) := & \frac{1}{m} \sum_{j=1}^m h_{i,j}(a_i),
\\
h(a) := & \frac{1}{n} (h_1(a_1), \cdots, h_n(a_n)),
\end{align*}
where the last term $h(a)$ is the concatenation of all $h_i$ for $i=1,\cdots,n$.
In Appendix , we provide the details of how (\ref{eq:min_max_general_form}) is specifically viewed as a stochastic problem.


As mentioned above while discussing the key challenge of the compositional structure in RO-TS (\ref{eq:ROTS}), conventional SGDA methods for Problem (\ref{eq:min_max_general_form}) require us to compute the full gradient of the compositional part $g(h_i(a_i))$, i.e., $\nabla h_i(a_i)^\top \nabla g(h_i(a_i))$, which involves {\it all} alignments in the case of RO-TS.

In contrast, SCAGDA only samples a constant number of $h_{i,j}(a_i)$ over $j$ (i.e., over a subset of alignments for RO-TS) and $\nabla h_{i,j}(a_i)$ ({\bf Line \ref{alg:line:random_compositional_elements}}).
Subsequently, SCAGDA employs a simple iterative {\em moving average} (MA) approach to accumulate $h_{i,j}(a_i)$ into $\omega^{k+1}_i$ at iteration $k$ for estimating $h_i(a_i)$ ({\bf Line \ref{alg:line:mv_compositional}}).
The key idea behind moving average method is to control the variance of the estimation for $h_i(a_i^{k+1})$ using a weighted average from the previous estimate $h_i(a_i^{k})$. 
Even though $\nabla g( \omega^{k+1}_i )$ is a biased estimation of $\nabla g(h_i(a_i^k))$, we can still use smoothness condition (introduced in Section \ref{section:theoretical_analysis} later) and bound the approximation error $\E[ \| \omega_i^{k} - h_i(a_i^{k-1}) \|^2 ]$ where $\omega_k := (\omega_1^k, ..., \omega_n^k)$ is the concatenation of all $\{ \omega_i^k \}_{i=1}^n$ at iteration $k$, as shown in Theorem \ref{theorem:convergence_approximation_error}.

Therefore, instantiation of SCAGDA for RO-TS does not require us to perform computation over {\em all} alignments contained in $h_i(a_i)$ for each time-series training sample, which leads to a more efficient algorithm with high scalability on large datasets. 
As shown in {\bf Line \ref{alg:line:primal_gradient}} and {\bf \ref{alg:line:dual_gradient}}, SCAGDA updates the primal variable $w$ and dual variable $a$ in an {\it alternating} scheme, which means that $w_{k+1}$ is updated based on $a_k$, while $a_{k+1}$ is updated based on $w_{k+1}$. This is different from SGDA, which updates $a_k$ based on $w_k$ instead. 
We instantiate SCAGDA for the proposed RO-TS framework as shown in Algorithm \ref{alg:pseudocode}. 
The primal variable update is provided in Line \ref{alg:pseudocode:line:update_primal}, and the dual variable update is provided in Line \ref{alg:pseudocode:line:update_dual}. 
In particular, Line \ref{alg:pseudocode:line:moving_average1} and \ref{alg:pseudocode:line:moving_average2} correspond to the moving average step for estimating $k_\GAK$ using randomly sampled alignment subset $\widehat{\mathcal A}_i^k$ for the $i$-th time-series training example.

\begin{algorithm}[t]
\caption{RO-TS Instantiation of SCAGDA}
\label{alg:pseudocode}
\textbf{Input}: A training set $\{(x, y) \in \mathcal{X} \times \mathcal Y\}^{n_{\text{train}}}$; mini-batch size $s$, deep neural network $f(w,x,y)$; learning rates $\eta_k$ and $\gamma_k$, loss function $l(\cdot)$, distance function $D(\cdot, \cdot)$.\\
\textbf{Output}: Classifier weights $w \in \Theta$
\begin{algorithmic}[1] 
\STATE Randomly initialize weights of the DNN classifier: $w_{0} \in \Theta$ \\
// vector of worst-case perturbations, one for each time-series
\STATE Initialize $a_{0} =0 $ and $\omega_0=0$ \\
// Multiple iterations of SCAGDA
\FOR{$k = 0, \cdots, K-1$}  
\STATE Randomly sample a mini-batch of data samples indexed by $\mathcal I_k$ s.t. $|\mathcal I_k|=s$

\STATE Compute the stochastic gradient w.r.t. $w_k$ \\ \hspace{2ex}$\mathcal{G}_{w,k} = \frac{1}{s} \sum_{i \in \mathcal I_k} \nabla_w l(f(w_k, x_i + a_i^k, y_i) $

\STATE Perform stochastic gradient descent on $w_k$\\ \hspace{2ex} $w_{k+1} = w_{k} - \eta_k \mathcal{G}_{w,k}$
\label{alg:pseudocode:line:update_primal}


\STATE Randomly sample a mini-batch of alignments indexed by $ \widehat{\mathcal A}_i^k$ for each data index $i \in \mathcal I_k$

\STATE Moving average for $i \in \mathcal I_k$\\ 
$\omega_i^{k+1}=(1-\beta)\omega_i^k + \beta \sum_{\pi \in \widehat{\mathcal A}_i^k} \exp(-d_{\pi}( x_i, x_i + a_i^k ) / \nu)$ 
\label{alg:pseudocode:line:moving_average1}

\label{alg:pseudocode:line:moving_average2}
\STATE $\omega_i^{k+1} = \omega_i^{k}$ for $i \notin \mathcal I_k$.

\STATE Compute
$\mathcal{G}_{a,k,i} = \nabla_a \ell(f(w_{k+1},x_i+a_i^k,y_i) - \sum_{\pi \in \widehat{\mathcal A}_i^k}\exp(- d_{\pi}(x_i, x_i + a_i^k) / \nu ) \cdot \nabla_a d_{\pi} (x_i, x_i + a_i^k) \cdot \frac{\lambda}{\omega^{k+1}_i \nu}$ for $i \in \mathcal I_k$


\STATE Perform stochastic gradient ascent over perturbations  \\  \hspace{2ex} $a_i^{k+1} = a_i^k + \gamma_k \mathcal{G}_{a,k,i} $
\label{alg:pseudocode:line:update_dual}

\ENDFOR
\STATE \textbf{return} weights of the learned DNN classifier, $w_K$
\end{algorithmic}
\end{algorithm}

In the next section, we show that our algorithm can converge to primal gap $P(w_K) - \min_w P(w) \leq \epsilon$ with iteration complexity $O(1/\epsilon^2)$, where $\epsilon$ is a pre-defined threshold. 
{\em To the best of our knowledge, this is the first optimization algorithm and convergence analysis for the famaily of compositional min-max optimization problems shown in (\ref{eq:min_max_general_form}).}

\section*{Theoretical Analysis}
\label{section:theoretical_analysis}

In this section, we present novel theoretical convergence analysis for SCAGDA algorithm. 
As mentioned in the previous section, for the problem (\ref{eq:min_max_general_form}), existing theoretical analysis of SGDA \cite{lin2020near,yan2020optimal}, stochastic alternating gradient descent ascent (SAGDA) \cite{yang2020global} require us to compute exact gradient of $g(h_i(a_i))$ at each iteration.
On the other hand, it is unclear if stochastic compositional alternating gradient algorithms for minimization problems \cite{wang2017stochastic,chen2020solving} can handle the complex min-max case. 

\vspace{1.0ex}

\noindent {\bf Summary of results.} We answer the following question: {\em can we establish convergence guarantee of our SCAGDA algorithm for nonconvex-nonconcave compositional min-max optimization problems?}

 Theorem \ref{theorem:convergence_rate} proves that SCAGDA shown in Algorithm \ref{algorithm:SCAGDA} converges to an $\epsilon$-primal gap in $O(\frac{1}{\epsilon^2})$ iterations. 
Theorem \ref{theorem:convergence_approximation_error} demonstrates the efficacy of the moving average strategy to approximate GAK based distance: the approximation error 
$\| \omega^K_i - h_i(a_i^{K-1}) ) \|^2$ 
is also bounded by $\epsilon$ in expectation when the $\epsilon$-primal gap is achieved.

\subsection*{Main Results}

The following commonly used assumptions are used in our analysis. 
Due to the space limit, definitions, proofs, and detailed constant dependencies can be found in Appendix .

\begin{assumption}\label{assumptions}
Suppose $\mu, L, C_g, C_h, L_g, L_h \geq 0$.
\\
(i)
$\phi(w, a)$ satisfies two side $\mu$-PL (Polyak-Lojasiewicz) condition:
\begin{align*}
\| \nabla_w \phi(w, a) \|^2 \geq 2 \mu ( \phi(w, a) - \min_{w'} f(w, a) )  ,
\\
\| \nabla_a \phi(w, a) \|^2 \geq 2 \mu ( \max_{a'} f(w, a) - \phi(w, a) )  .
\end{align*}

\noindent
(ii) $\phi(w, a)$ is $L$-smooth in $w$ for fixed $a$.

\noindent
(iii) $\phi(w, a)$ is $L$-smooth in $a_i$ for fixed $w$.

\noindent
(iv) $g$ (resp. $h$) is $C_g$ (resp. $C_h$)-Lipschitz continuous.

\noindent
(v) $g$ (resp. $h$) is $L_g$ (resp. $L_h$)-smooth. 

\noindent
(vi) $\exists$ $\sigma > 0$ s.t. \; $\E[\| \nabla_w \phi_i(w, a_i) - \nabla_w \phi(w, a) \|^2] \leq \sigma^2$, 

\noindent
$\E[\| \nabla_a f_i(w, a_i) - \nabla_a f_i(w, a_i) \|^2] \leq \sigma^2$,
$\E[ \| h_{i, j}(a_i) - h(a) \|^2 ] \leq \sigma^2$, and
$\E[ \| \nabla h_{i, j}(a_i) - \nabla h(a) \|^2 ] \leq \sigma^2$
\end{assumption}

\noindent We present our main results for SCAGDA below. 

\begin{theorem}\label{theorem:convergence_rate}
Suppose Assumption \ref{assumptions} holds.
In Algorithm \ref{algorithm:SCAGDA}, set $\eta_k = \eta = O(1 / \epsilon^2)$, $\gamma_k = \gamma = O(1 / L^2 )$ and $\beta = \sqrt{18\mu\eta}$.
After running Algorithm \ref{algorithm:SCAGDA} for $K$ iterations where $K = \tilde O( 1 / \epsilon^2 )$ ($\tilde O$ hides logarithmic factor), we have
\begin{align*}
&
\E[P(w_K) - P^*] 
+ \frac{1}{8} \E[ P(w_K) - \phi(w_K, a_K) ]
\\
&
+ \Big( \frac{ 4 C_h^4 L_g^4 \eta_K }{ \mu^5 } \Big)^{1/2} \E [ \| \omega_K - h(a_{K-1}) \|^2 ]
\leq 
\epsilon 
\end{align*}
\end{theorem}

\noindent {\bf Remark 1.}
The above theorem gives us two critical observations of the behavior of SCAGDA. {\bf (1)} After running $K$ iterations of SCAGDA, the primal gap $P(w_{K+1}) - P^*$ converges to $\epsilon$ in expectation, since all terms in the left hand side of the inequality are non-negative. 
This result shows that SCAGDA is able to effectively solve the compositional min-max optimization problem shown in Equation (\ref{eq:min_max_general_form}).
{\bf (2)} The required iteration complexity of SCAGDA is $O(1/\epsilon^2)$. 
To put this result in perspective, we compare it with related theoretical results. 
The rate for nonconvex-nonconcave min-max problem without compositional structure is shown to be $O(1/\epsilon)$ \cite{yang2020global}. 
However, this improvement requires unbiased estimation (or exact value) of the gradient and computing the exact $g(h_i(a_i))$ at each iteration. 
Our iteration complexity is in the same order of that for \cite{chen2020solving}, whose convergence result is $O(1/\epsilon^2)$ for nonconvex compositional {\it minimization} problems instead of {\it min-max} ones. 
The difference is that their convergence metric is the average squared norm of gradients, while ours is for the primal gap. Importantly, {\em this is the first result on convergence rate for stochastic compositional min-max problems}.

\begin{theorem}
\label{theorem:convergence_approximation_error}
After $K = \tilde O( 1 / \epsilon^2 )$ iterations of Algorithm \ref{algorithm:SCAGDA}, we have:
$
\E [ \| \omega^K_i - h_i(a^{K-1}_i) \|^2 ]
\leq 
O(\epsilon) .
$
\end{theorem}

\vspace{0.5ex}

\noindent {\bf Remark 2.}
The above result shows that as SCAGDA algorithm is executed, the approximation error of $\| \omega^K_i - h_i(a^{K-1}_i) \|^2$ converges to $\epsilon$ in the expectation as it is achieving the $\epsilon$-primal gap.
For the condition numbers, we always have $L \geq \mu$. 
In practice, we usually set the accuracy level $\epsilon$ to a very small value, so the condition $\epsilon \leq O(L^3/\mu^2)$ will generally hold. This result provides strong theoretical support that if we apply SCAGDA to optimize our RO-TS problem in (\ref{eq:ROTS}), it is able to approximate $k_{\GAK}$ on-the-fly, where we only need a constant number of alignments, rather than {\it all} possible alignments for computing $k_{\GAK}$ in each iteration of SCAGDA. 
When we have $\epsilon$-primal gap, we also achieve $\epsilon$-accurate estimation of $k_{\GAK}$ at the same time.

\section*{Related Work}

Prior work on robustness of DNNs is mostly focused on image/text domains; and can be classified into two categories.

\vspace{1.0ex}

\noindent {\bf Adversarial training} employs augmented data such as adversarial examples \cite{kolter2018materials,wang2019deep} and input perturbations.  Methods to create adversarial examples include general attacks such as Carlini \& Wagner attack \cite{carlini2017towards}, boundary attack \cite{brendel2017decision}, and universal attacks \cite{moosavi2017universal}. Recent work regularizes adversarial example generation methods to obey intrinsic properties of images \cite{laidlaw2019functional,xiao2018spatially,hosseini2017limitation}. There are also specific  adversarial methods for NLP domain \cite{samanta2017towards,gao2018black}. There is little to no prior work on adversarial techniques for time-series domain. Fawaz et al. \cite{fawaz2019adversarial} employed the standard Fast Gradient Sign method \cite{kurakin2016adversarial} to create adversarial noise for time-series. 
Network distillation was also employed to train a student model for creating adversarial attacks \citep{karim2020adversarial}. This method is severely limited: it can generate adversarial examples for only a small number of target labels and cannot guarantee generation of adversarial example for every input.

\vspace{1.0ex}

\noindent {\bf Training via explicit loss function} employ an explicit loss function to capture the robustness criteria and optimize it. Stability training \cite{ZhengSLG16,LiCWC19} for images is based on the criteria that similar inputs should produce similar DNN outputs. Adversarial training can be interpreted as min-max optimization, where a hand-designed optimizer such as projected gradient descent is employed to (approximately) solve inner maximization. \cite{xiong2020improved} train a neural network to guide the optimizer. 
Since characteristics of time-series (e.g., fast-pace oscillations, sharp peaks) are different from images/text, $L_p$ distance based methods are not suitable for time-series domain. 

\vspace{0.5ex}

In summary, there is no prior work\footnote{In a concurrent work, \cite{TSA-STAT} proposed an adversarial framework for time-series domain using statistical features and also provided robustness certificates.} to train robust DNNs for time-series domain in a principled manner. This paper precisely fills this important gap in our scientific knowledge.

\section*{Experiments and Results}

We present experimental evaluation of RO-TS on real-world time-series benchmarks and compare with prior methods. \label{sec:experiment}

\subsection*{Experimental Setup}

\vspace{0.5ex}

\noindent \textbf{Datasets.} We employed diverse univariate and multi-variate time-series benchmark datasets from the UCR repository \cite{ucrdata}. Table \ref{tab:dsnames} describes the details of representative datasets for which we show the results (due to space limits) noting that our overall findings were similar on other datasets from the UCR repo. We employ the standard training/validation/testing splits for these datasets.

\begin{table}[b]
\centering
\begin{tabular}{ccc}  
\textbf{Name}  & \textbf{Classes} & \textbf{Input Size} \textbf{($C\times T$) }  \\ \hline
ECG200 & 2& 1$\times$97 \\ BME &3 &1$\times$129 \\
ECG5000 & 5& 1$\times$141 \\ MoteStrain &2 &1$\times$85 \\
SyntheticControl &6 & 1$\times$61 \\ RacketSports & 4&6$\times$30 \\
ArticularyWR & 25& 9$\times$144  \\ ERing & 6&4$\times$65 \\
FingerMovements &2 & 28$\times$50\\
\end{tabular}
\caption{Description of different datasets.}
\label{tab:dsnames}
\end{table}

\vspace{0.5ex}

\noindent \textbf{Algorithmic setup and baselines.} We employ a 1D-CNN architecture \cite{bai2018empirical} as the deep model for our evaluation. The details of the neural architecture are provided in the Appendix. We ran RO-TS algorithm shown in Appendix for a maximum of 500 iterations to train robust models. To estimate GAK distance within RO-TS, we employed 15 percent of the total alignments noting that larger sample sizes didn't improve the optimization accuracy and increased the training time. We also employ adversarial training to create models using baseline attacks that are not specific to image domain for comparison: Fast Gradient Sign method (FGS) \citep{kurakin2016adversarial} that was used by \citet{fawaz2019adversarial} and Projected Gradient Descent (PGD)\citep{madry2017towards}. 
We also compare RO-TS against stability training (STN) \cite{zheng2016improving}. 

\vspace{0.5ex}

\noindent \textbf{Evaluation metrics.} We evaluate the robustness of created models using different attack strategies on the testing data. The prediction accuracy of each model (via ground-truth labels of time-series) is used as the metric. To ensure robustness, DNN models should be least sensitive to different types of perturbations over original time-series signals. We measure the accuracy of each DNN model against: {\em 1) Adversarial noise} is introduced by FGS and PGD baseline attacks; and {\em 2) Gaussian noise} $\sim~\mathcal{N}(0, \Sigma)$ that may naturally occur to perturb time-series. The covariance matrix $\Sigma$ diagonal elements (i.e., variances) are all equal to $\sigma$. DNNs are considered robust if they are successful in maintaining their accuracy performance against such noises.

\subsection*{Results and Discussion}
\begin{figure}[!h]
    \centering
        \begin{minipage}{\linewidth}
        \begin{minipage}{\linewidth}
            \centering
            Random Noise
        \end{minipage}
        \begin{minipage}{.33\linewidth}
                \centering                \includegraphics[width=\linewidth]{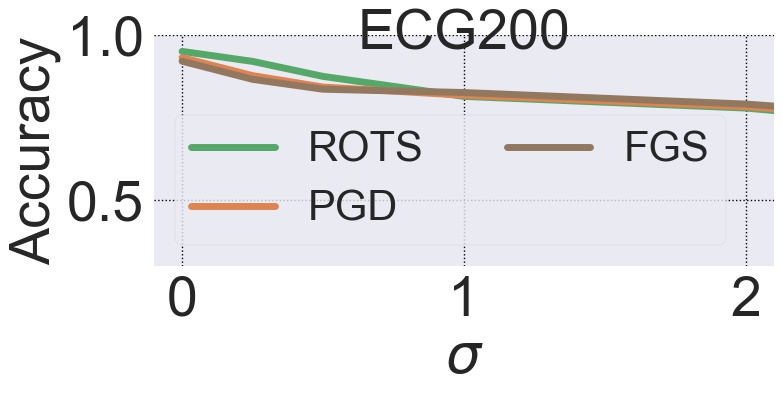}
            \end{minipage}%
        \begin{minipage}{.33\linewidth}
                \centering
                \includegraphics[width=\linewidth]{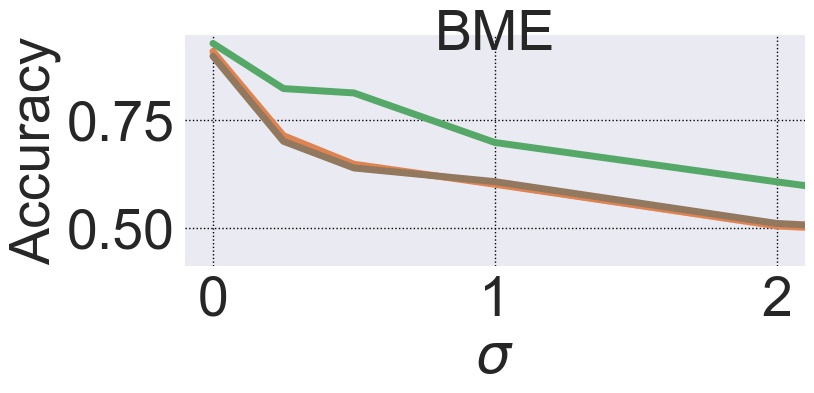}
            \end{minipage}%
        \begin{minipage}{.33\linewidth}
                \centering
                \includegraphics[width=\linewidth]{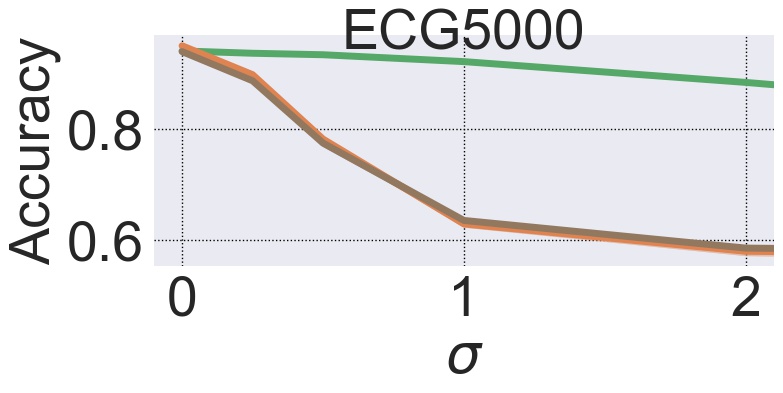}
            \end{minipage}
        \begin{minipage}{.33\linewidth}
                \centering
                \includegraphics[width=\linewidth]{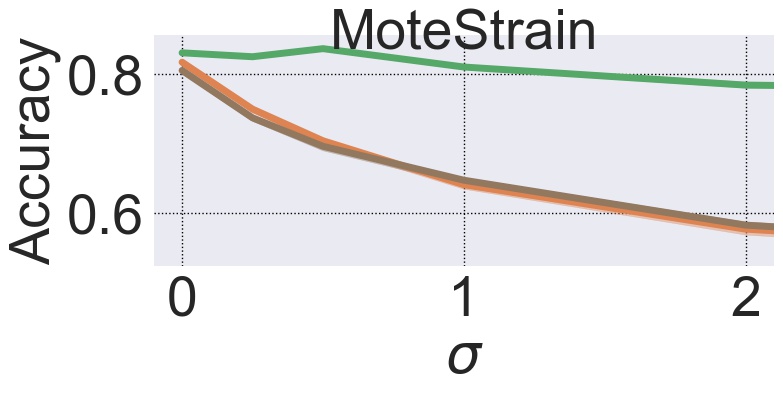}
            \end{minipage}%
        \begin{minipage}{.33\linewidth}
                \centering
                \includegraphics[width=\linewidth, height=.6in]{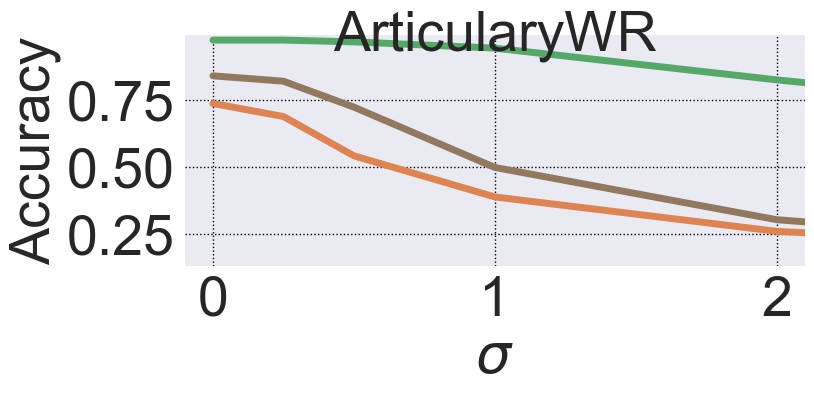}
            \end{minipage}%
        \begin{minipage}{.33\linewidth}
                \centering
                \includegraphics[width=\linewidth]{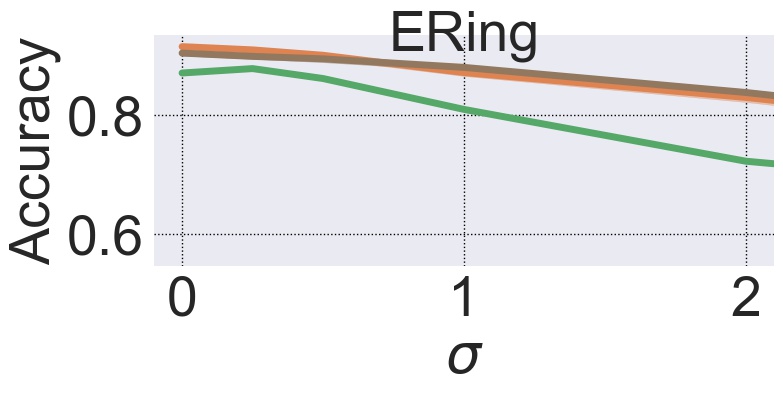}
            \end{minipage}
        \begin{minipage}{.33\linewidth}
                \centering
                \includegraphics[width=\linewidth]{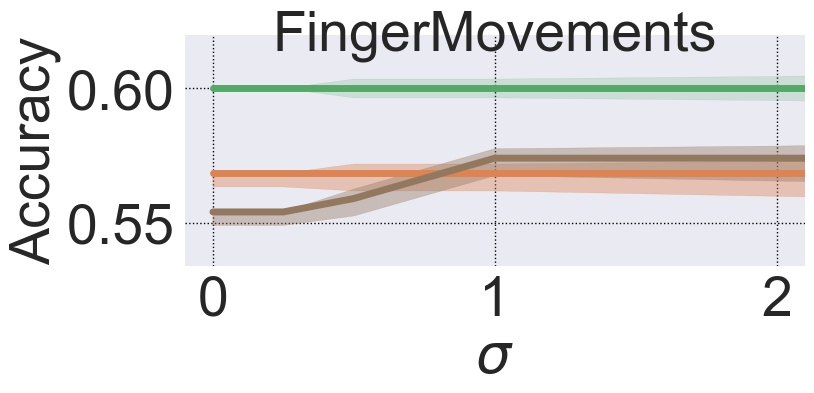}
            \end{minipage}%
        \begin{minipage}{.33\linewidth}
                \centering
                \includegraphics[width=\linewidth]{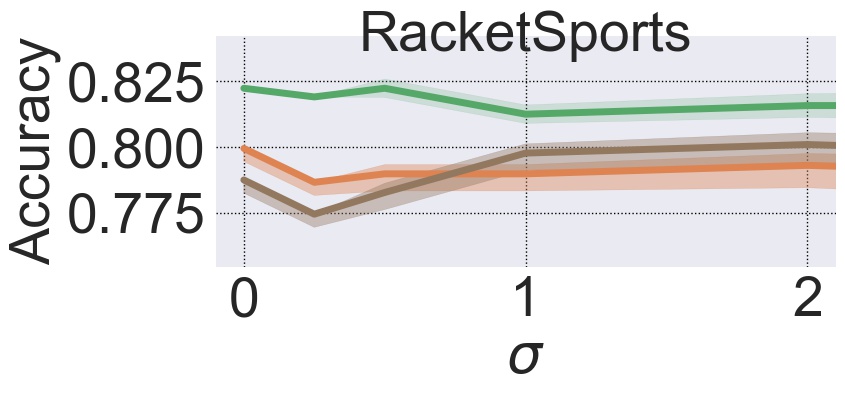}
            \end{minipage}%
        \begin{minipage}{.33\linewidth}
                \centering
                \includegraphics[width=\linewidth]{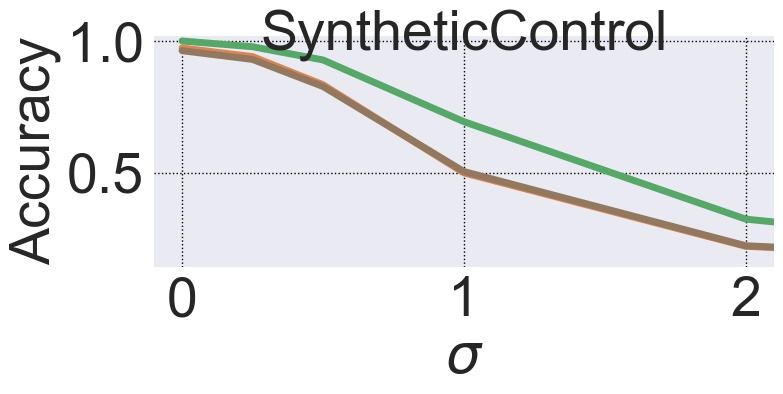}
            \end{minipage}
        \begin{minipage}{\linewidth}
            \centering
            Adversarial Noise
        \end{minipage}
    \end{minipage}
    \begin{minipage}{\linewidth}
        \begin{minipage}{.33\linewidth}
                \centering
                \includegraphics[width=\linewidth]{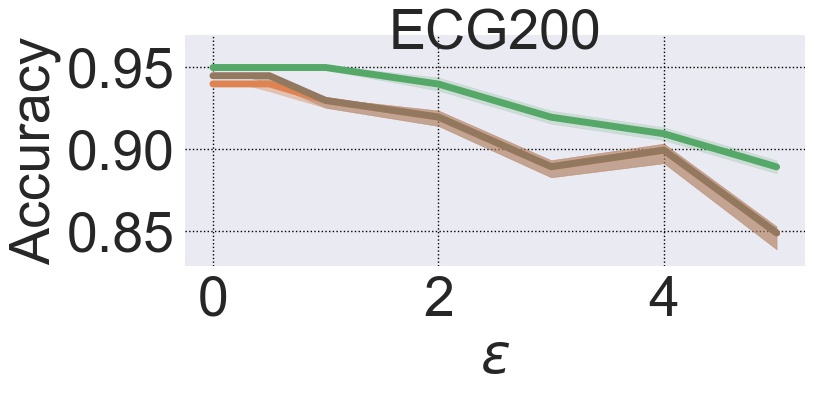}
            \end{minipage}%
        \begin{minipage}{.33\linewidth}
                \centering
                \includegraphics[width=\linewidth]{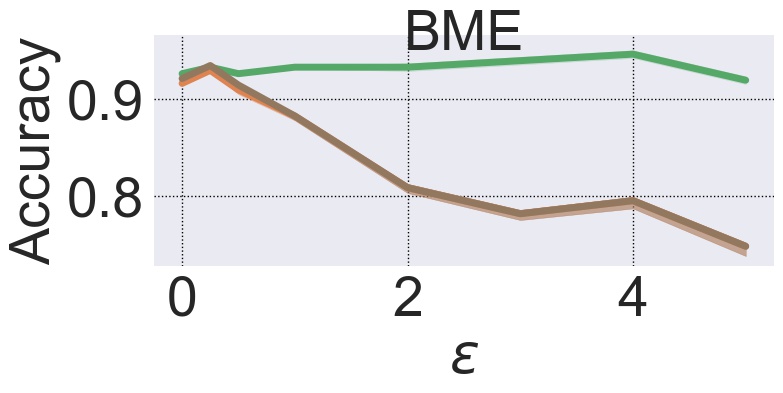}
            \end{minipage}%
        \begin{minipage}{.33\linewidth}
                \centering
                \includegraphics[width=\linewidth]{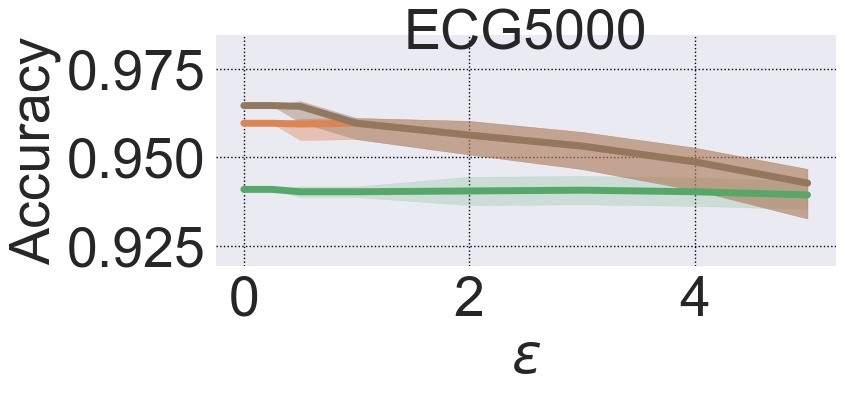}
            \end{minipage}
        \begin{minipage}{.33\linewidth}
                \centering
                \includegraphics[width=\linewidth]{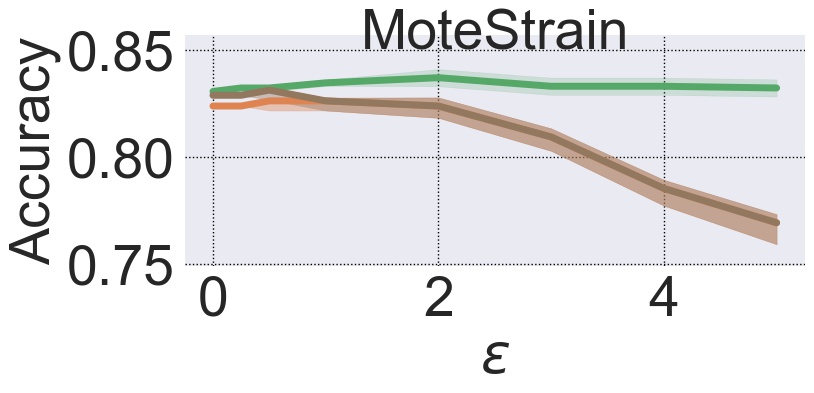}
            \end{minipage}%
        \begin{minipage}{.33\linewidth}
                \centering
                \includegraphics[width=\linewidth, height=.6in]{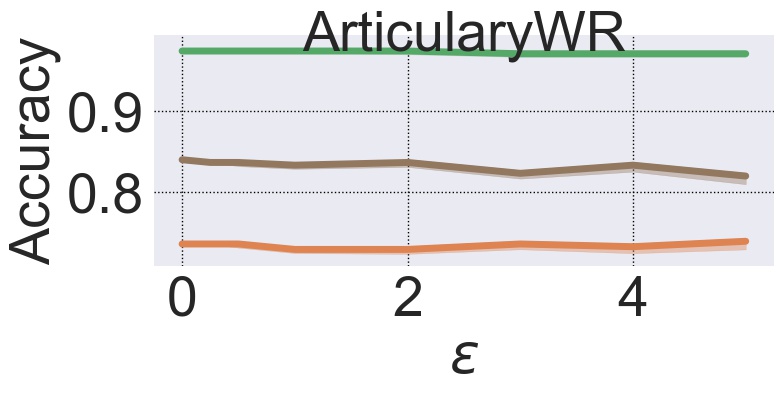}
            \end{minipage}%
        \begin{minipage}{.33\linewidth}
                \centering
                \includegraphics[width=\linewidth]{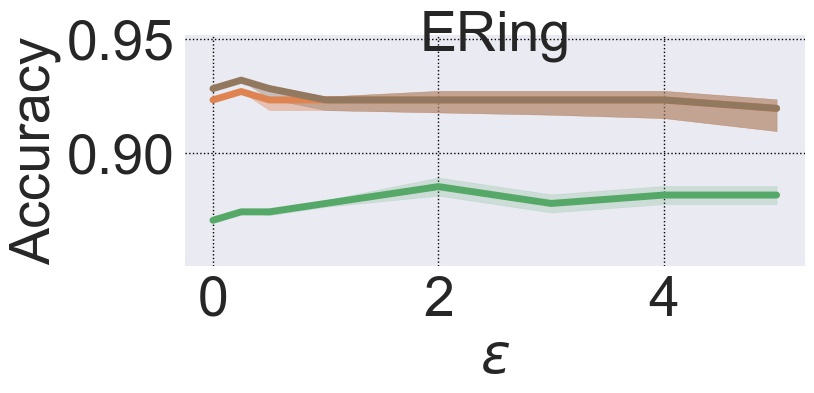}
            \end{minipage}
        \begin{minipage}{.33\linewidth}
                \centering
                \includegraphics[width=\linewidth]{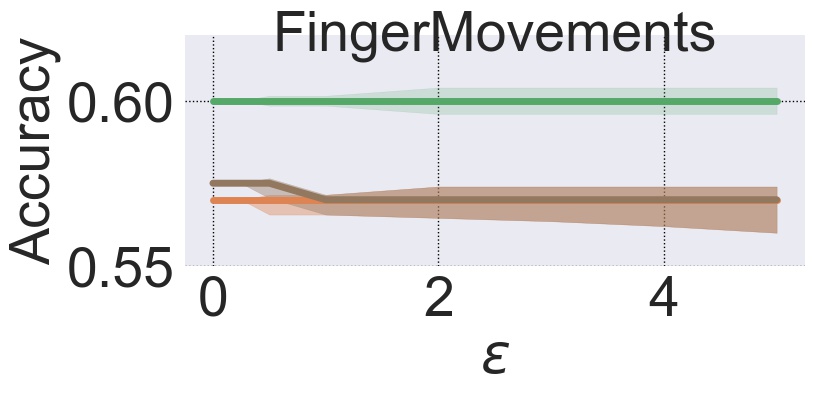}
            \end{minipage}%
        \begin{minipage}{.33\linewidth}
                \centering
                \includegraphics[width=\linewidth]{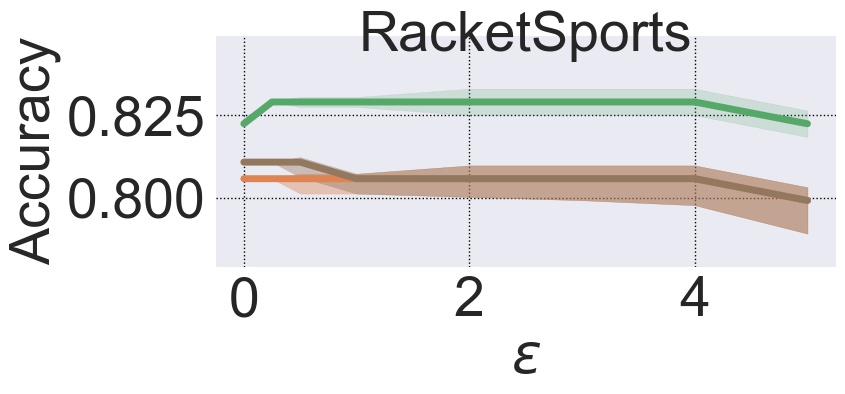}
            \end{minipage}%
        \begin{minipage}{.33\linewidth}
                \centering
                \includegraphics[width=\linewidth]{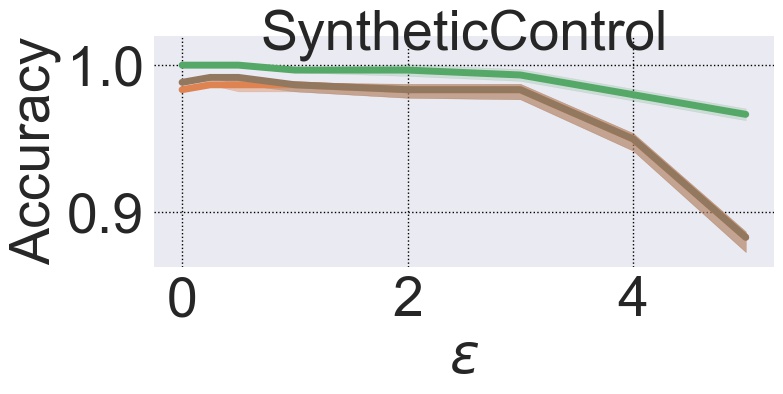}
            \end{minipage}
    \end{minipage}
    
\caption{Comparison of RO-TS algorithm vs. adversarial training algorithm using baseline FGS and PGD attacks.} 
\label{fig:advtrn}
\end{figure}%
\begin{figure}[!h]
    \centering
        \begin{minipage}{\linewidth}
        \begin{minipage}{\linewidth}
            \centering
            Random Noise
        \end{minipage}
        \begin{minipage}{.33\linewidth}
                \centering
                \includegraphics[width=\linewidth]{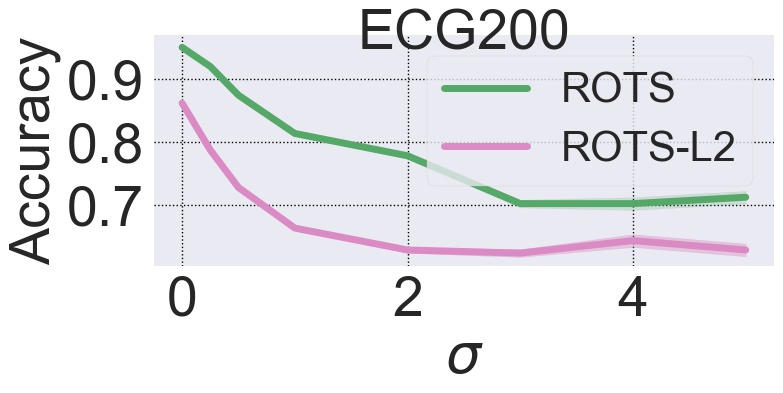}
            \end{minipage}%
        \begin{minipage}{.33\linewidth}
                \centering
                \includegraphics[width=\linewidth]{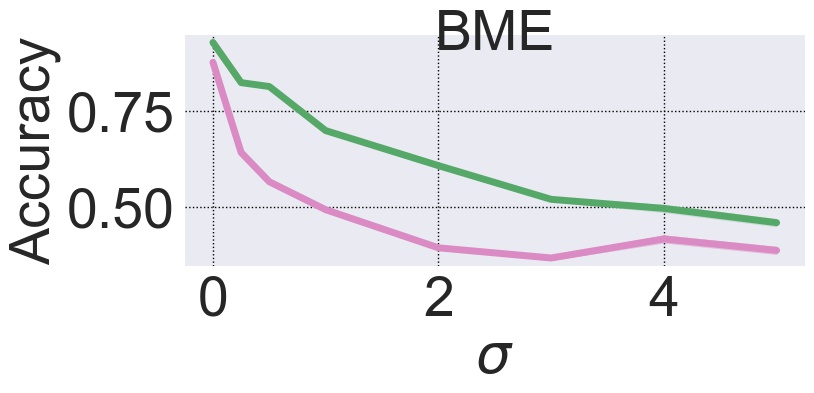}
            \end{minipage}%
        \begin{minipage}{.33\linewidth}
                \centering
                \includegraphics[width=\linewidth]{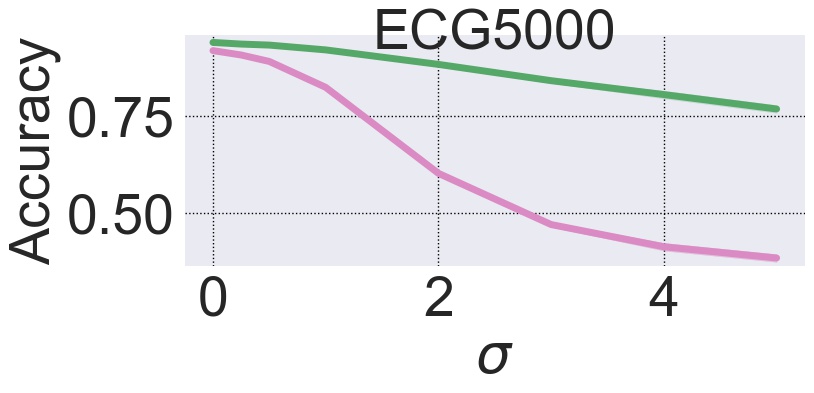}
            \end{minipage}
        \begin{minipage}{.33\linewidth}
                \centering
                \includegraphics[width=\linewidth]{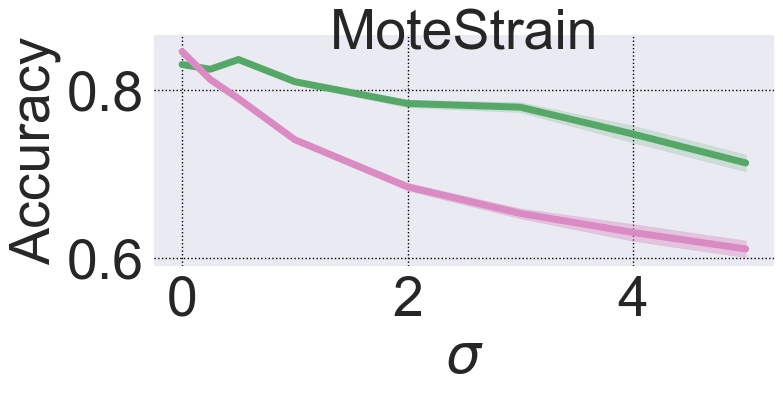}
            \end{minipage}%
        \begin{minipage}{.33\linewidth}
                \centering
                \includegraphics[width=\linewidth, height=.45in]{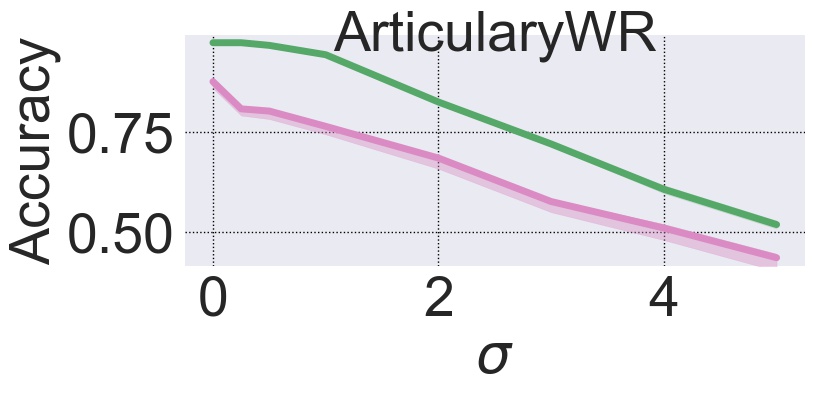}
            \end{minipage}%
        \begin{minipage}{.33\linewidth}
                \centering
                \includegraphics[width=\linewidth]{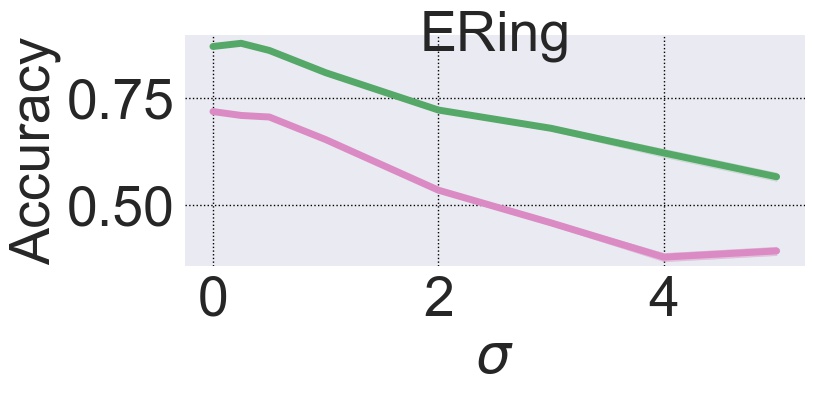}
            \end{minipage}
        \begin{minipage}{.33\linewidth}
                \centering
                \includegraphics[width=\linewidth]{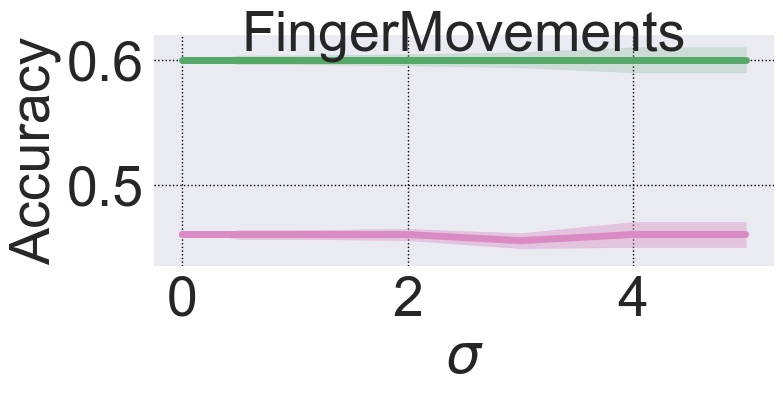}
            \end{minipage}%
        \begin{minipage}{.33\linewidth}
                \centering
                \includegraphics[width=\linewidth]{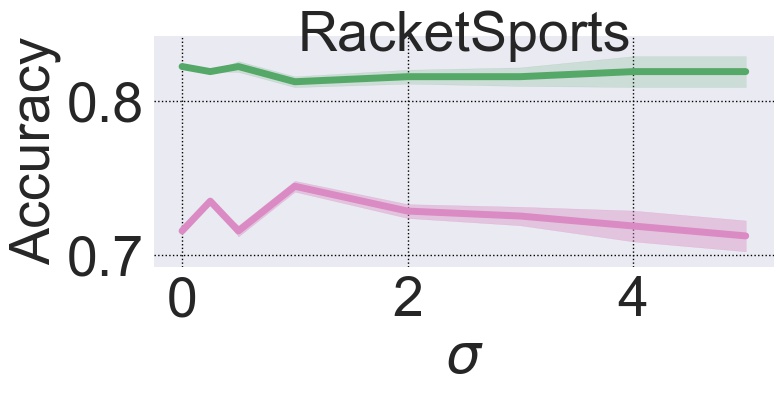}
            \end{minipage}%
        \begin{minipage}{.33\linewidth}
                \centering
                \includegraphics[width=\linewidth]{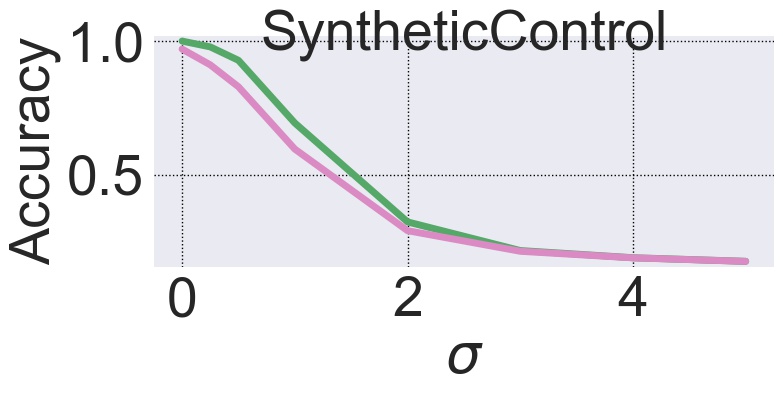}
            \end{minipage}
        \begin{minipage}{\linewidth}
            \centering
            Adversarial Noise
        \end{minipage}
    \end{minipage}
    \begin{minipage}{\linewidth}
        \begin{minipage}{.33\linewidth}
                \centering
                \includegraphics[width=\linewidth]{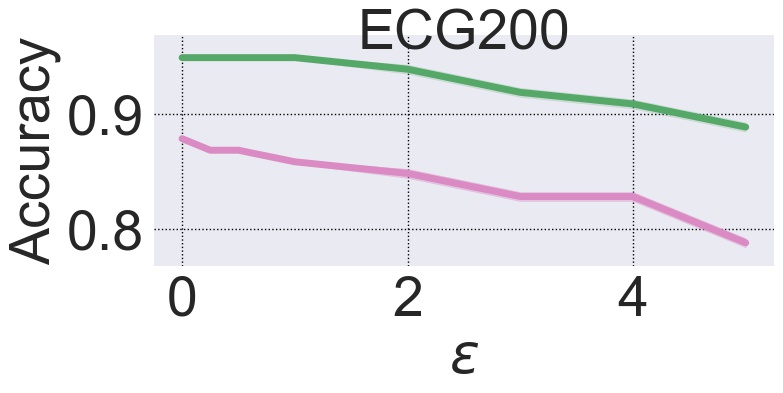}
            \end{minipage}%
        \begin{minipage}{.33\linewidth}
                \centering
                \includegraphics[width=\linewidth]{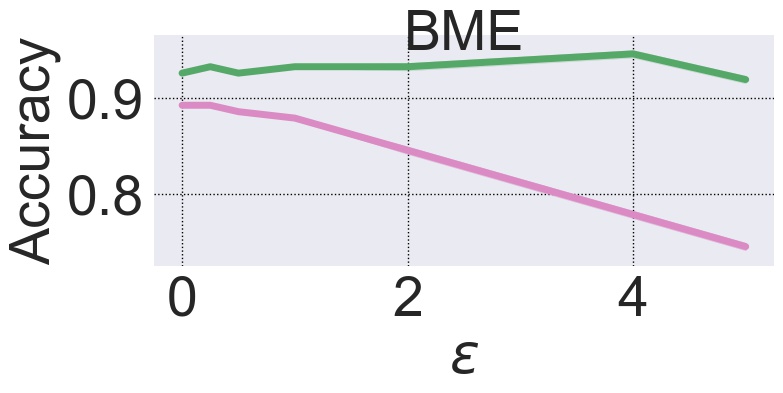}
            \end{minipage}%
        \begin{minipage}{.33\linewidth}
                \centering
                \includegraphics[width=\linewidth]{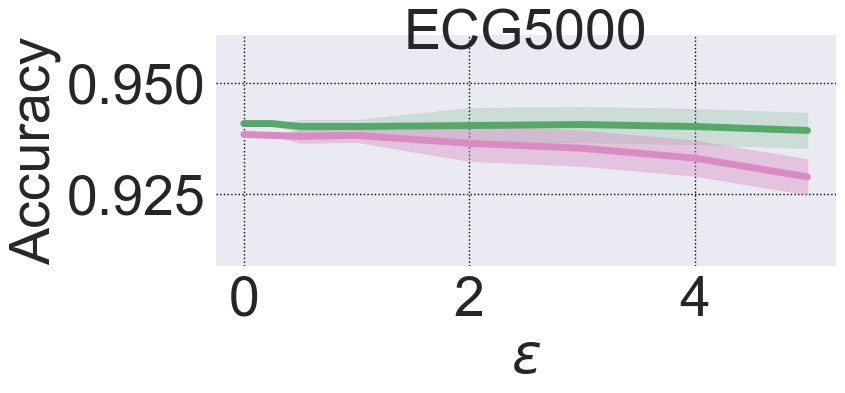}
            \end{minipage}
        \begin{minipage}{.33\linewidth}
                \centering
                \includegraphics[width=\linewidth]{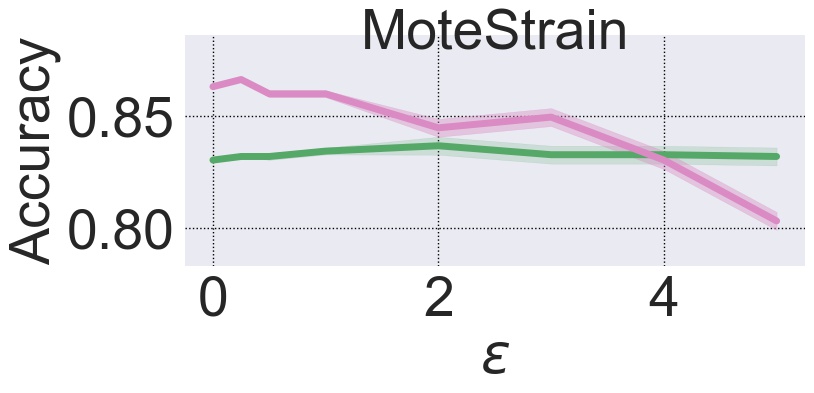}
            \end{minipage}%
        \begin{minipage}{.33\linewidth}
                \centering
                \includegraphics[width=\linewidth, height=.45in]{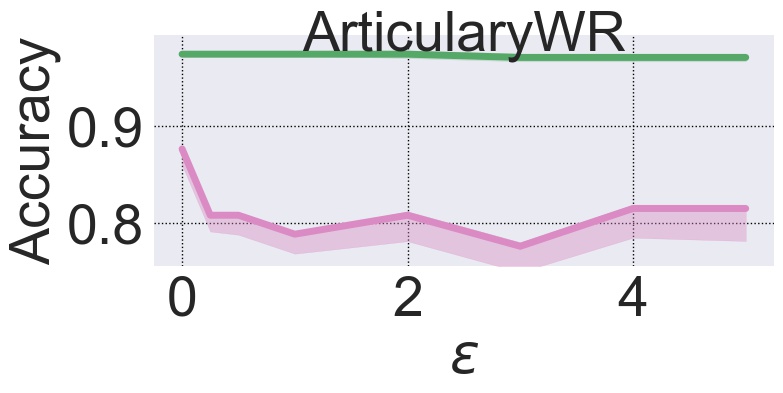}
            \end{minipage}%
        \begin{minipage}{.33\linewidth}
                \centering
                \includegraphics[width=\linewidth]{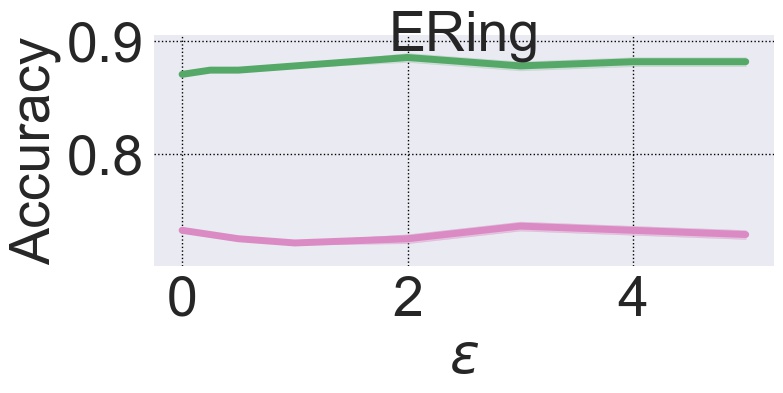}
            \end{minipage}
        \begin{minipage}{.33\linewidth}
                \centering
                \includegraphics[width=\linewidth]{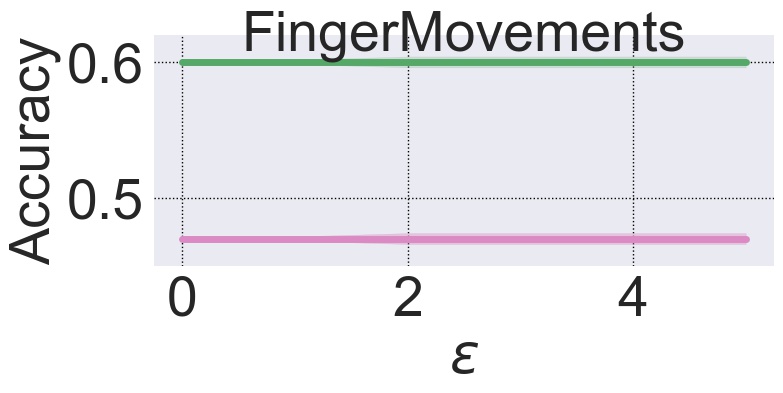}
            \end{minipage}%
        \begin{minipage}{.33\linewidth}
                \centering
                \includegraphics[width=\linewidth]{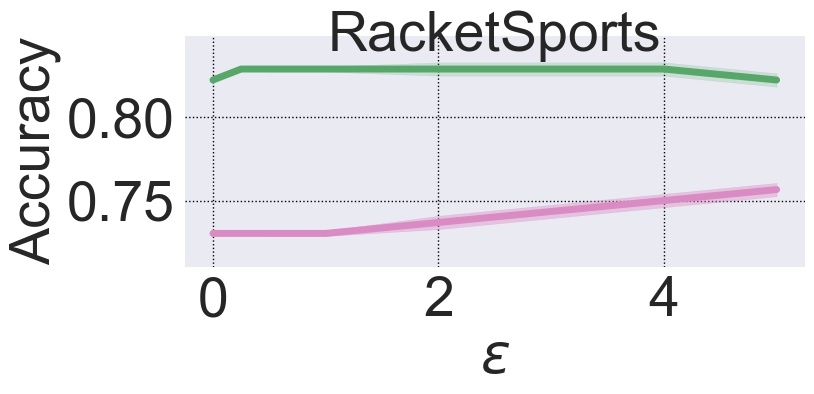}
            \end{minipage}%
        \begin{minipage}{.33\linewidth}
                \centering
                \includegraphics[width=\linewidth]{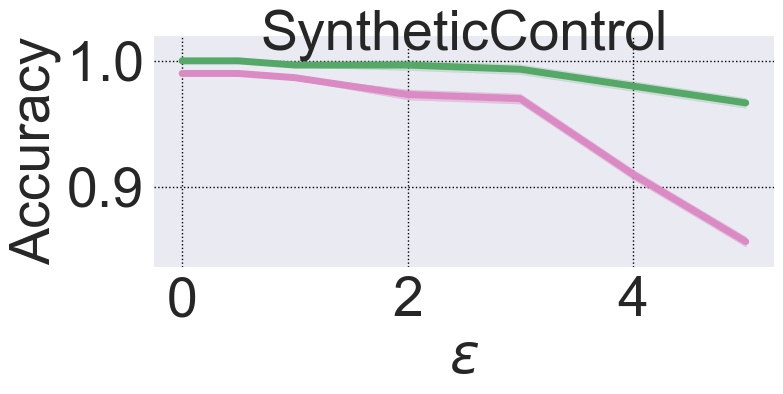}
            \end{minipage}
    \end{minipage}
\caption{Comparison of RO-TS algorithm using GAK distance ($k_{\GAK}$) vs. RO-TS using Euclidean distance ($L_2$).} 
\label{fig:gakvsl2}
\end{figure}%
\begin{figure}[!h]
    \centering
        \begin{minipage}{\linewidth}
        \begin{minipage}{\linewidth}
            \centering
            Random Noise
        \end{minipage}
        \begin{minipage}{.33\linewidth}
                \centering
                \includegraphics[width=\linewidth]{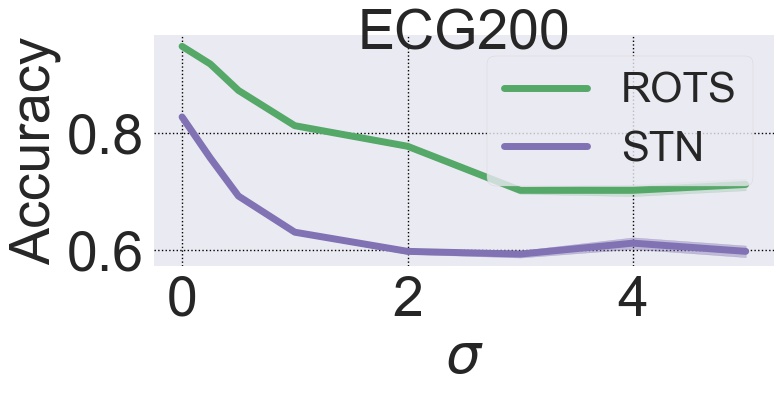}
            \end{minipage}%
        \begin{minipage}{.33\linewidth}
                \centering
                \includegraphics[width=\linewidth]{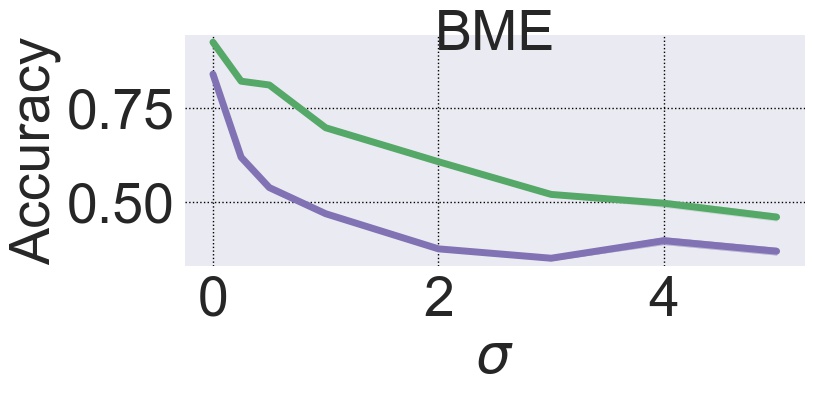}
            \end{minipage}%
        \begin{minipage}{.33\linewidth}
                \centering
                \includegraphics[width=\linewidth]{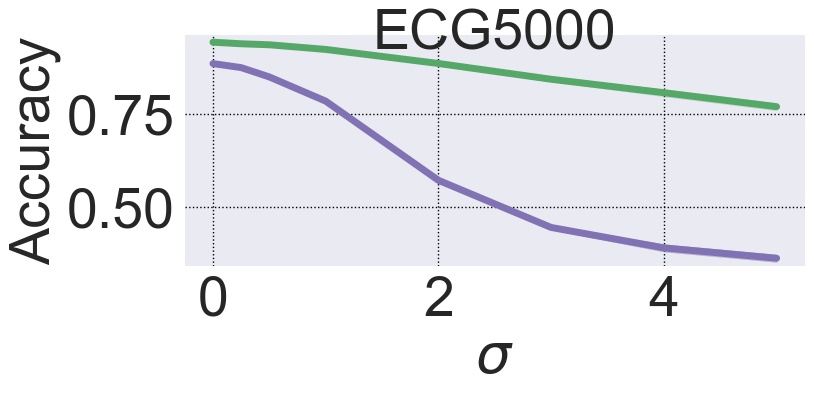}
            \end{minipage}
        \begin{minipage}{.33\linewidth}
                \centering
                \includegraphics[width=\linewidth]{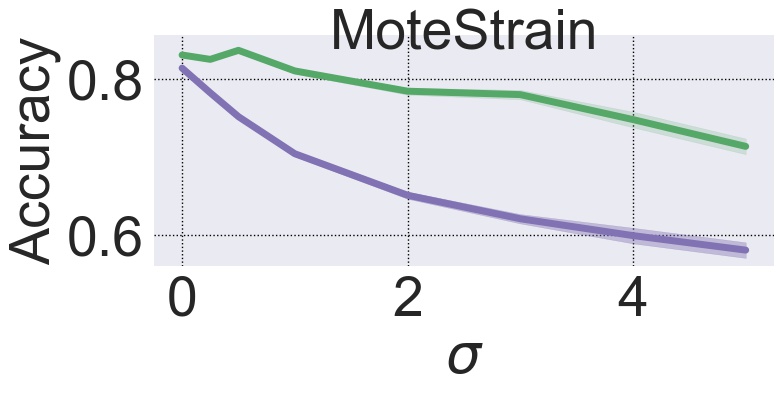}
            \end{minipage}%
        \begin{minipage}{.33\linewidth}
                \centering
                \includegraphics[width=\linewidth, height=.45in]{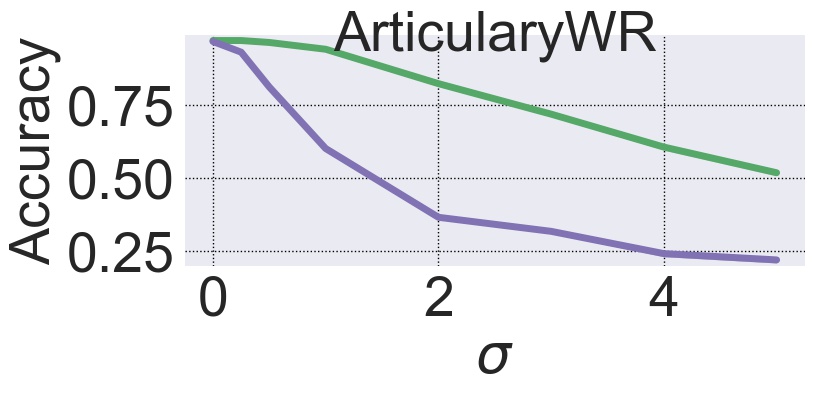}
            \end{minipage}%
        \begin{minipage}{.33\linewidth}
                \centering
                \includegraphics[width=\linewidth]{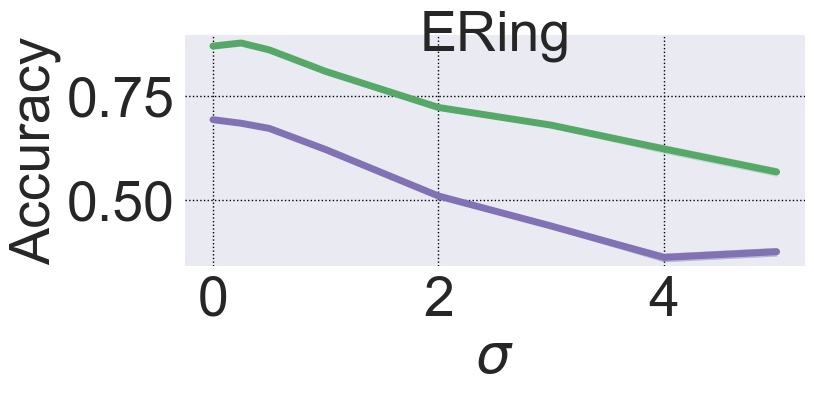}
            \end{minipage}
        \begin{minipage}{.33\linewidth}
                \centering
                \includegraphics[width=\linewidth]{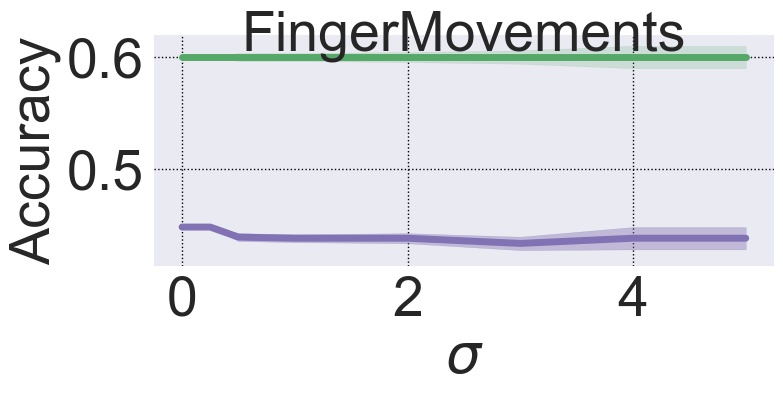}
            \end{minipage}%
        \begin{minipage}{.33\linewidth}
                \centering
                \includegraphics[width=\linewidth]{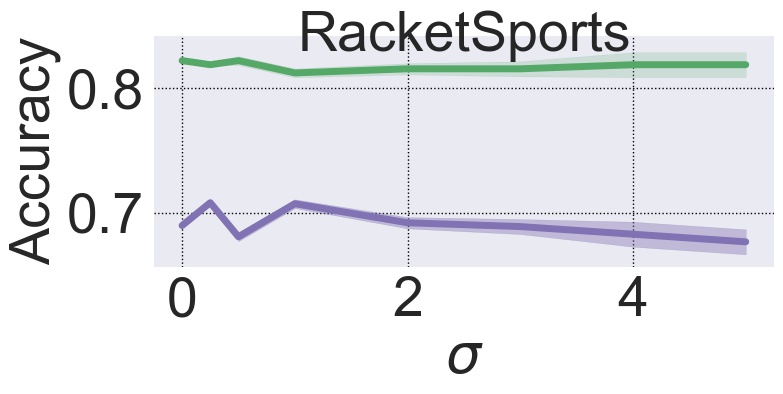}
            \end{minipage}%
        \begin{minipage}{.33\linewidth}
                \centering
                \includegraphics[width=\linewidth]{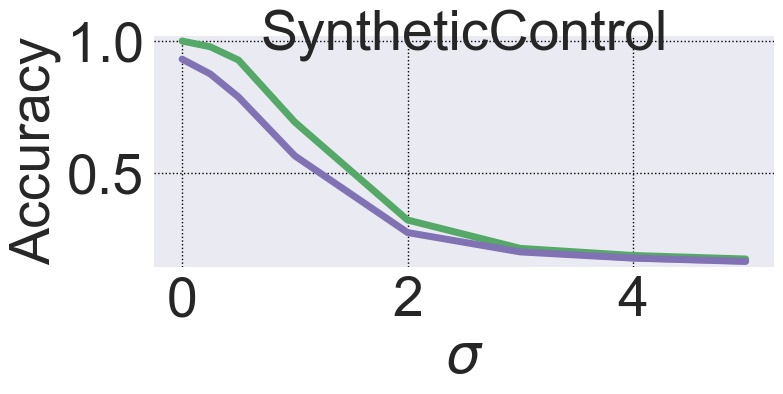}
            \end{minipage}
        \begin{minipage}{\linewidth}
            \centering
            Adversarial Noise
        \end{minipage}
    \end{minipage}
    \begin{minipage}{\linewidth}
        \begin{minipage}{.33\linewidth}
                \centering
                \includegraphics[width=\linewidth]{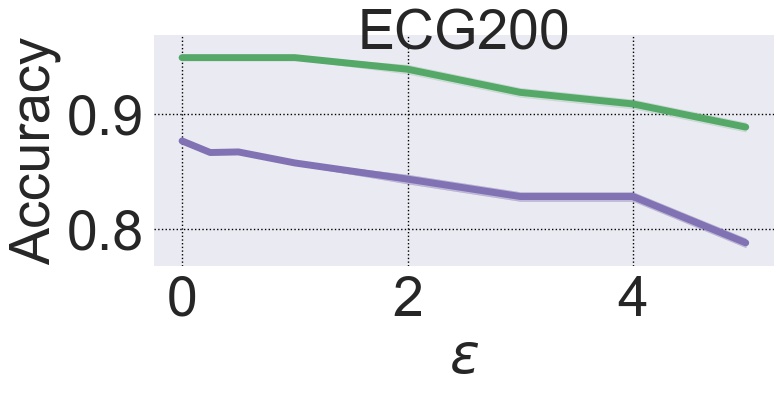}
            \end{minipage}%
        \begin{minipage}{.33\linewidth}
                \centering
                \includegraphics[width=\linewidth]{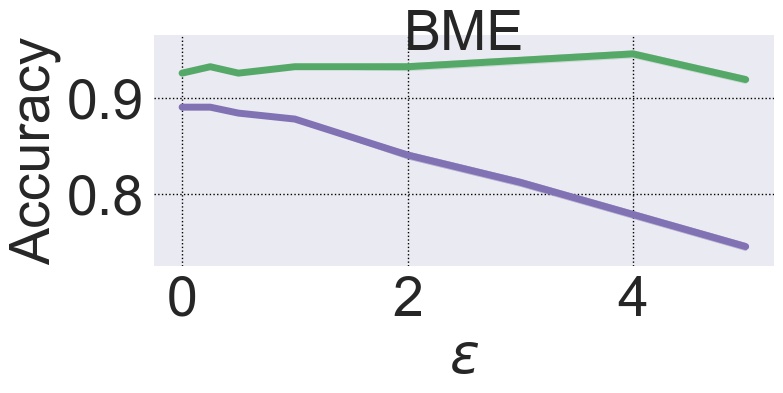}
            \end{minipage}%
        \begin{minipage}{.33\linewidth}
                \centering
                \includegraphics[width=\linewidth]{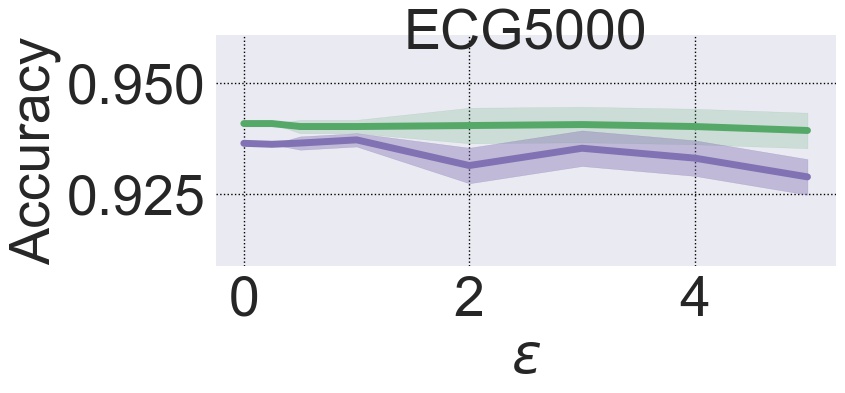}
            \end{minipage}
        \begin{minipage}{.33\linewidth}
                \centering
                \includegraphics[width=\linewidth]{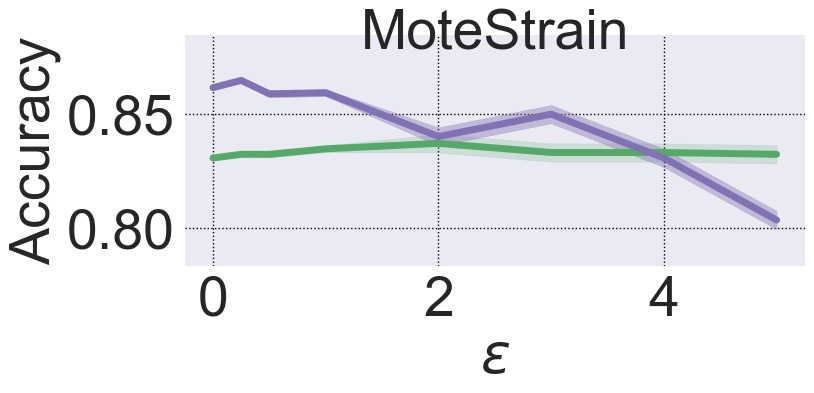}
            \end{minipage}%
        \begin{minipage}{.33\linewidth}
                \centering
                \includegraphics[width=\linewidth, height=.45in]{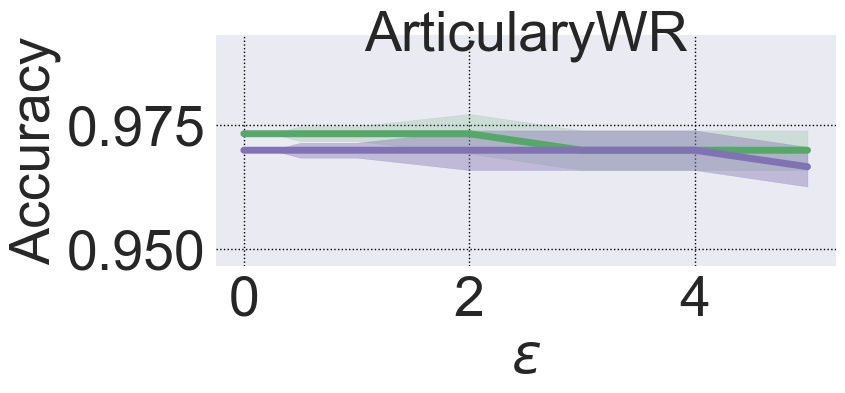}
            \end{minipage}%
        \begin{minipage}{.33\linewidth}
                \centering
                \includegraphics[width=\linewidth]{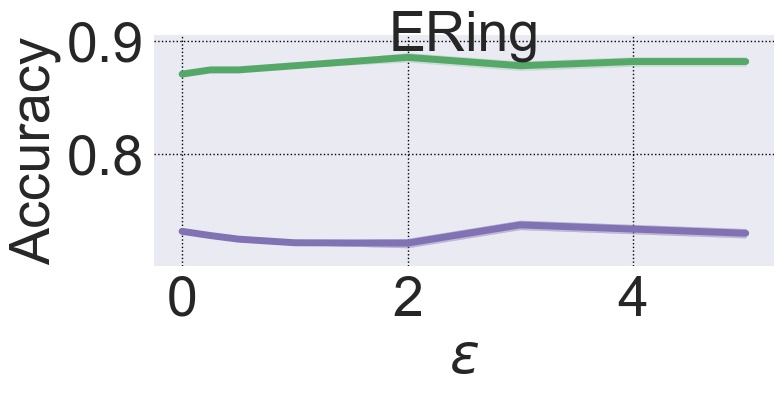}
            \end{minipage}
        \begin{minipage}{.33\linewidth}
                \centering
                \includegraphics[width=\linewidth]{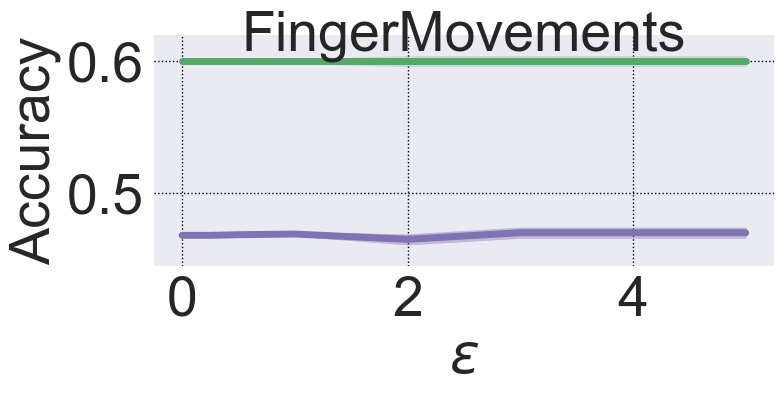}
            \end{minipage}%
        \begin{minipage}{.33\linewidth}
                \centering
                \includegraphics[width=\linewidth]{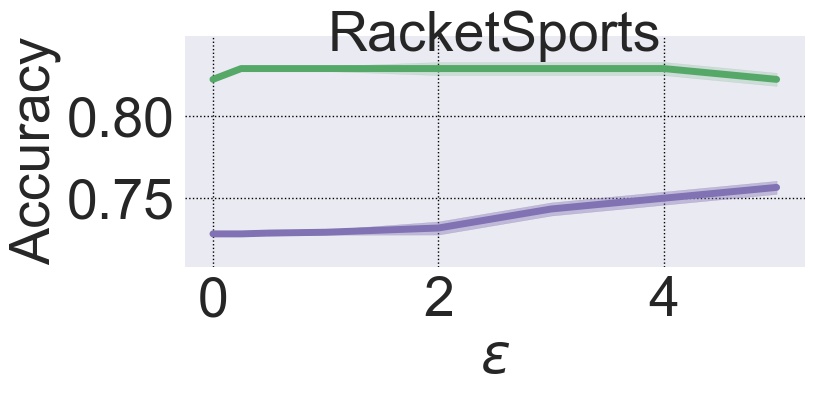}
            \end{minipage}%
        \begin{minipage}{.33\linewidth}
                \centering
                \includegraphics[width=\linewidth]{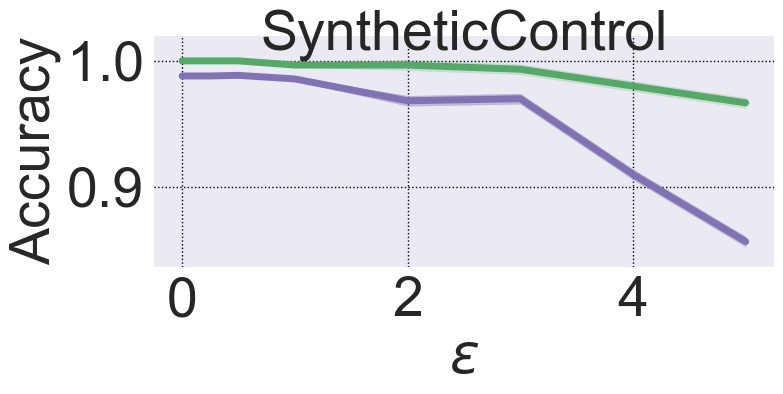}
            \end{minipage}
    \end{minipage}
\caption{Comparison of RO-TS vs. stability training (STN).}
\label{fig:stn}
\end{figure}

\vspace{0.5ex}

\noindent \textbf{RO-TS vs. adversarial training.} One of our key hypothesis is that Euclidean distance-based perturbations do not capture the appropriate notion of invariance for time-series domain to improve the robustness of the learned model. We show that using baseline attacks to create augmented data for adversarial training does not create robust models. From Figure \ref{fig:advtrn}, we can observe that models from our RO-TS algorithm achieve significantly higher accuracy than the baselines. For example, on MoteStrain dataset, RO-TS has a steady performance against both types of noises, unlike the baselines. On the other datasets, we can clearly observe that in most cases, RO-TS outperforms the baselines. We conclude that adversarial training using prior methods and attack strategies is not as effective as our RO-TS method, where we perform explicit primal-dual optimization to create robust models.

\vspace{0.5ex}

\noindent \textbf{RO-TS vs. RO-TS with L2 distance.} We want to demonstrate that choosing the right distance metric to compute similarity between time-series signals is critical to create robust models. Therefore, we compare models created by RO-TS by using two different distance metrics: 1) The standard Euclidean distance $\|\cdot\|_2$ used in image domains and prior work; and 2) Using the GAK distance $D_{\GAK}$. 
From Figure \ref{fig:gakvsl2}, we can clearly see that the GAK distance is able to explore the time-seris input space better to improve robustness. The Euclidean distance either  performs significantly worse than GAK (e.g., on ECG5000, ERing, and RacketSports datasets) or performs comparably to GAK (e.g., on SyntheticControl or ArticularyWR datasets). This experiment concludes that GAK is a suitable distance metric for time-series domain. 

\vspace{0.5ex}

\noindent \textbf{RO-TS vs. stability training.} Unlike adversarial training, stability training (STN) employs the below loss function \cite{zheng2016improving} to introduce stability to the deep model. 
\begin{equation*}
    Loss_{STN} = L_0(x, \theta) + 0.01\times L_{stability}(x, x', \theta)
\end{equation*}
where $x$ is the original input, $x'$ is a perturbed version of $x$ using additive Gaussian noise $\sim~\mathcal{N}(0, 0.04^2)$, $L_0$ is the 
cross-entropy loss, and $L_{stability}$ relies on KL-divergence. We experimentally demonstrate that 
RO-TS formulation is more suitable than STN for creating robust DNNs for time-serirs domain. Figure \ref{fig:stn} shows a comparison between DNNs trained using STN and RO-TS. We observe that for most datasets, RO-TS creates significantly more robust DNNs when compared to STN for both types of perturbations.
RO-TS algorithm is specifically designed for time-series domain by making appropriate design choices, whereas STN is designed for image domain. Hence, RO-TS allows us to create more robust DNNs for time-series domain.

\vspace{0.5ex}

\begin{figure}[t]
    \centering
    \includegraphics[width=.7\linewidth]{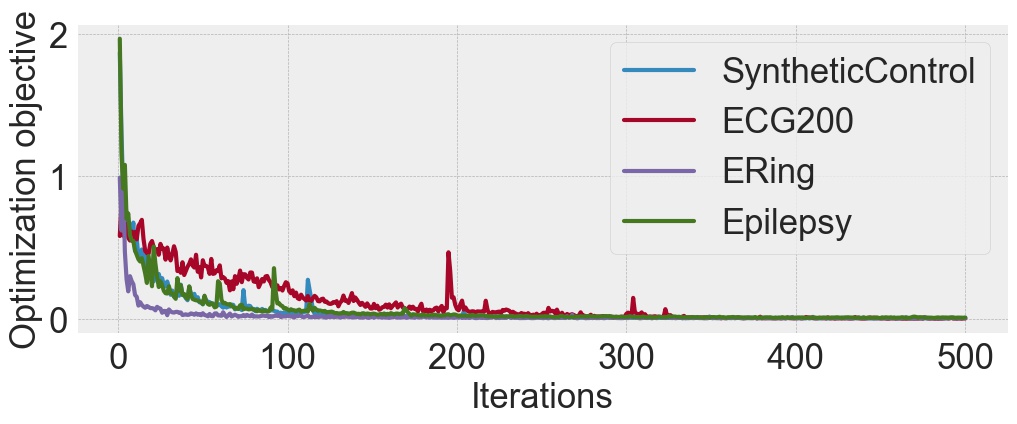}
    \caption{Empirical convergence of RO-TS algorithm.}
    \label{fig:conv}
\end{figure}

\begin{figure}[t]
    \centering
    \begin{minipage}{.49\linewidth}
        \centering
        \includegraphics[width=\linewidth]{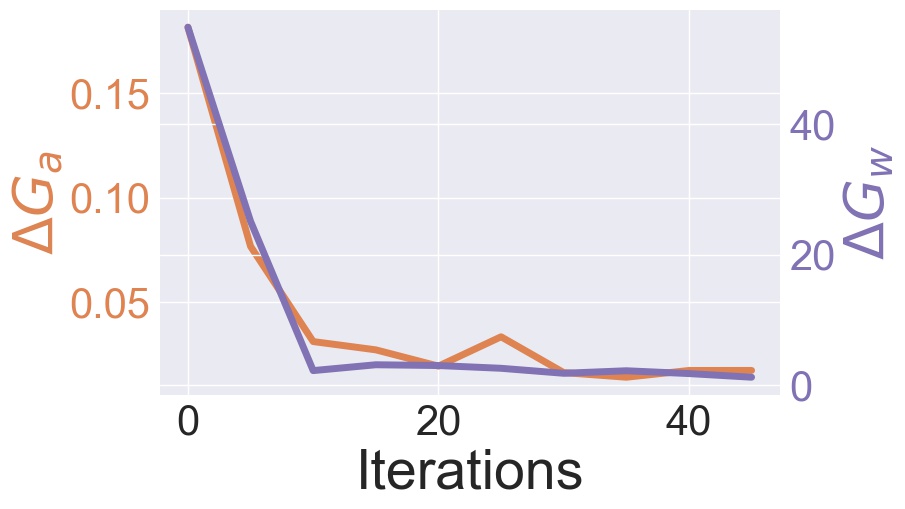}
    \end{minipage}
    \begin{minipage}{.49\linewidth}
        \centering
        \includegraphics[width=\linewidth]{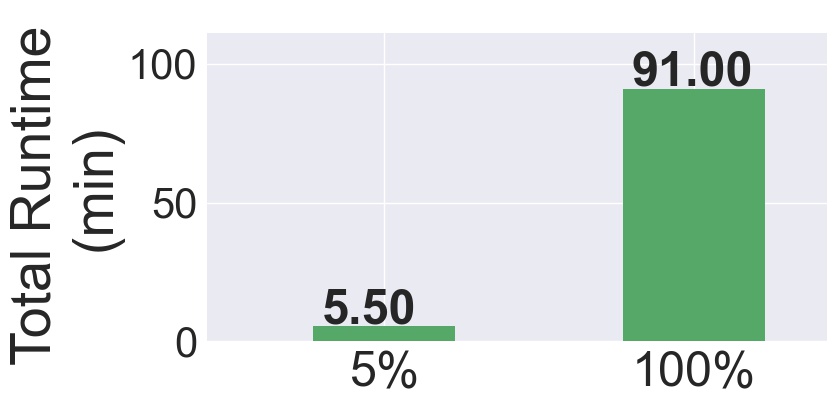}
    \end{minipage}
    \caption{The accuracy gap in the gradients over weights $\Delta G_W$ and over perturbations $\Delta G_a$ using 5\% of alignments and GAK using all alignments for RO-TS training on ERing (Left) and the comparison of the computational runtime between both settings of RO-TS (Right).}
    \label{fig:gradient}
\end{figure}

\noindent \textbf{Empirical convergence.} We demonstrate the efficiency of RO-TS algorithm by observing the empirical rate of convergence. Figure \ref{fig:conv} shows the optimization objective over iterations on some representative datasets noting that we observe similar patterns on other datasets.
We can observe that RO-TS converges roughly before 150 iterations for most datasets.
Figure \ref{fig:gradient} shows the accuracy gap results and the computational runtime when comparing RO-TS with sampled alignments and original GAK (i.e., all alignment paths). The results clearly match with our theoretical analysis that accuracy gap decreases over training iterations leading to convergence. We conclude from these results that RO-TS converges quickly in practice and supports our theoretical analysis.


\section*{Conclusions}
We introduced the RO-TS algorithm to train robust deep neural networks (DNNs) for time-series domain. The training problem was formulated as a min-max optimization problem to reason about the worst-case risk in a small neighborhood defined by the global alignment kernel (GAK) based distance. 
Our proposed stochastic compositional alternating gradient descent and ascent (SCAGDA) algorithm carefully leverages the structure of the optimization problem to solve it efficiently. Our theoretical and empirical analysis showed that RO-TS and SCAGDA are effective in creating more robust DNNs over prior methods and GAK based distance is better suited for time-series over the Euclidean distance.

\newpage

\section*{Acknowledgments} This research is supported in part by the AgAID AI Institute for Agriculture Decision Support, supported by the National Science Foundation and United States Department of Agriculture - National Institute of Food and Agriculture award \#2021-67021-35344.

\bibliography{references}

\newpage
\onecolumn
\appendix
\section{Experimental Details}
\label{appendix:section:experiment}

We employed the standard benchmark training, validation, and test split on the UCR datasets. Each experiment was repeated 10 times and the average has been reported. The shaded region in the graphs depicts the minimum and maximum value achieved within those experiments. We implemented our proposed RO-TS algorithm using TensorFlow and the baselines using the CleverHans library \cite{papernot2018cleverhans}. 

\vspace{1.0ex}

\noindent {\bf DNN Architecture.} We employed a 1D-CNN architecture for all our experiments (both RO-TS and baselines). The architecture as follows: \{C:100; K:5; C:50; K:5; P:4; R:200; R:100\} where C: Convolutional layers, K: kernel size, P: max-pooling kernel size, and R: rectified linear unit layer.

\vspace{1.0ex}

\noindent {\bf RO-TS.} We provide the pseudo-code of RO-TS to train robust DNNs in Algorithm \ref{alg:pseudocode}. 
The $\nu$ hyper-parameter is estimated as the median distance of different points observed in different time-series of the training set. 
It is then scaled by the square root of the median length of time-series in the training set \cite{cuturi2011fast}. 
For all our experiments reported in Section \ref{sec:experiment}, we employed $\lambda=10\mathrm{e}^{-2}$ to prevent exploding gradient problems and a fixed $\beta \in [0,1]$ value based on validation experiments to compute the moving average for GAK distance. We employed a mini-batch size $s \in [10,25]$ in all our experiments. The source code of RO-TS algorithms is available at \href{https://github.com/tahabelkhouja/Robust-Training-for-Time-Series.git}{Robust-Training-for-Time-Series GitHub}

\vspace{1.0ex}

\noindent {\bf Baselines.} To create the PGD and FGS baselines, we use adversarial examples created using perturbation factors $\epsilon <$ 1. This bound is used for two main reasons. First, larger perturbations significantly degrade the overall performance of adversarial training. Second, we want to avoid the risk of leaking label information \citep{madry2017towards}. Additionally, the STN baseline has been implemented as described in the original paper \cite{zheng2016improving}.  The adversarial training using the baseline attacks employ the same architecture and use Adam optimizer.

\vspace{1.0ex}

\noindent {\bf GAK distance.} For the computation of $k_{\GAK}(x, x')$ for time-series signals $x$ and $x'$: {\em 1) Defining the set of paths $\pi$:} Every path $\pi$ uses a set of two coordinates $\pi_1(i)$ and $\pi_2(i)$. We randomly sample set of paths $\{\pi\}$  for GAK computation such that $\|\pi_1(i)-\pi_2(i)\|< \frac{T}{2}$. This procedure avoids clearly bad alignments, where few time-steps from the first signal are matched with many time-steps from the second signal. 
{\em 2) Fixing the value of $\nu$:} As $\nu$ is the divisor of $d_\pi( x, x' )$, we compute the pairwise distance $d_{\pi_{diag}}(\cdot, \cdot)$ between all pairs of time-series signals from the training data and set $\nu$ as the median of all the different distances. $\pi_{diag}$ is the alignment path where we have a point-to-point match ($\|\pi_1(i)-\pi_2(i)\|=0$)

\section{Proofs of Technical Lemmas}
\label{sec:proofs}

Before we prove our results, we first present the detailed definitions for our assumptions.

\begin{definition}\label{def:lipschitz_continuous}
If a function $f(x)$ is $C$-Lipschitz continuous, then we have
\begin{align*}
| f(x_1) - f(x_2) | 
\leq 
C \| x_1 - x_2 \|  \text{ and } \| \nabla f(x) \| \leq C.
\end{align*}
\end{definition}

\begin{definition}\label{def:smooth}
If a function $f(x)$ is $L$-smooth, then its gradient is $L$-Lipschitz continuous:
\begin{align*}
\| \nabla f(x_1) - \nabla f(x_2) \| \leq L \| x_1 - x_2 \|  ,
\end{align*}
and equivalently
\begin{align*}
f(x_1) - f(x_2)
\leq 
\langle \nabla f(x_2), x_1 - x_2 \rangle + \frac{L}{2} \| x_1 - x_2 \|^2  .
\end{align*}
\end{definition}

\begin{definition}\label{def:pl_condition}
If a function $f(x)$ has a non-empty solution set and satisfies $\mu$-PL condition, then we have
\begin{align*}
\| \nabla f(w) \|^2 \geq 2 \mu ( f(x) - \min_{x'} f(x') )  .
\end{align*}
\end{definition}

We now present two important lemmas that can be directly derived from our Assumption \ref{assumptions}, whose proofs have been done by previous studies.

\begin{lemma} (Lemma A.2 and A.3 in \cite{yang2020global})
If $\phi$ is $L$-smooth and satisfies two side $\mu$-PL condition, then $P(w)$ also satisfies $\mu$-PL conditions, and $P(w)$ is $L_P$-smooth with $L_P=L+L^2/\mu \leq 2L^2/\mu$ when $L \geq 1$ and $\mu \leq 1$.
\end{lemma}

\begin{lemma}\label{lemma:PL_EB}
(\cite{karimi2016linear}) If $f(x)$ is $L$-smooth and satisfies $\mu$-PL condition, then it also satisfies $\mu$-error bound condition $\| \nabla f(x) \| \geq \mu \| x_p - x \|$ where $x_p$ is the projection of $x$ onto the optimal set.
Also it satisfies $\mu$-quadratic growth condition $f(x) - \min_x f(x) \geq \frac{\mu}{2} \| x_p - x \|^2$.
Conversely, if $f(x)$ is $L$-smooth and satisfies $mu$-error bound, then it satisfies $\frac{\mu}{L}$-PL condition.
\end{lemma}

From the next subsection, we give the detailed proofs for our Theorem \ref{theorem:convergence_rate} and \ref{theorem:convergence_approximation_error}
The rate of convergence rate is a quantity that represents how quickly the sequence approaches its limit. In our convergence analysis for our proposed optimization algorithm, this reveals the upper bound for the primal gap, i.e., $P(w) - P(w^*)$ where $w^* \in \arg\min_{w \in \Theta} P(w)$, as a function of $K$, the total number of iterations.

\subsection{ Proof of Theorem \ref{theorem:convergence_rate} }
Before we can start proof of the main theorem, we need the following three technical lemmas. 

\begin{lemma}
Suppose Assumption \ref{assumptions} holds.
Assume $P$ is $L_P$-smooth.
For iteration $k$ of Algorithm \ref{algorithm:SCAGDA}, we have
\label{lemma:primal_gap_recursion}
\begin{align}\label{eq:primal_gap_recursion}
\E [ P(w_{k+1}) - P^* ]
\leq &
P(w_k) - P^*
- \frac{\eta_k}{2} \| \nabla P(w_k) \|^2
+ \frac{L \eta_k^2 \sigma^2}{2}  
\nonumber\\
&
+ \frac{\eta_k}{2} \| \nabla_w \phi(w_k, a_k) - \nabla P(w_k) \|^2
\end{align}
\end{lemma}


\begin{lemma}
\label{lemma:auxiliary_recursion}
Suppose Assumption \ref{assumptions} hold.
For iteration $k$ of Algorithm \ref{algorithm:SCAGDA}, we have
\begin{align}\label{eq:auxiliary_recursion}
&
\E [ P(w_{k+1}) - \phi(w_{k+1}, a_{k+1}) ]
\nonumber\\
&
\leq 
(1 - \frac{\mu\gamma_k}{2}) \E[ P(w_k) - \phi(w_k, a_k) ]
\nonumber\\
&
+ (1 - \frac{\mu\gamma_k}{2}) ( \frac{\eta_k}{2} + 2\eta_k + L \eta_k^2 ) \| \nabla P(w_k) - \nabla \phi_w^k \|^2 
\nonumber\\
&
+ (1 - \frac{\mu\gamma_k}{2}) ( - \frac{\eta_k}{2} + 2\eta_k + L \eta_k^2) \| \nabla P(w_k) \|^2
\nonumber\\
&
+ 2 \gamma_k C_h^2 L_g^2 \E[ \| \omega_{k+1} - h(a_k) \|^2 ]
+ L \sigma^2 ( \eta_k^2 + \gamma^2 (1 + 2 C_g^2) )
\end{align}
\end{lemma}

\begin{lemma}
\label{lemma:compositional_recursion}
(Lemma 2 of \cite{wang2017stochastic})
\begin{align}\label{eq:compositional_recursion}
\E [ \| \omega_{k+1} - h(a_k) \|^2 ]
\leq &
(1 - \beta_k) \| \omega_k - h(a_{k-1}) \|^2
\nonumber\\
&
+ \frac{ C_h^2 \| a_k - a_{k-1} \|^2 }{\beta_k}
+ 2 \beta_k^2 \sigma^2 
\end{align}
\end{lemma}

We now have the three important inequalities that are required for our proof for Theorem \ref{theorem:convergence_rate}. The key idea is to construct a Lyapunov function that combines the above three inequalities together, and derive the convergence for this Lyapunov function.
Let 
\begin{align*}
L_{k+1} 
:= &
\E[P(w_{k+1}) - P^*] + \frac{1}{8} \E[ P(w_{k+1}) - \phi(w_{k+1}, a_k) ] 
\\
&
+ \Big( \frac{ 4 C_h^4 L_g^4 L^4 \eta_k }{\mu^5} \Big)^{1/2} \E[ \| \omega_{k+1} - h(a_k) \|^2]  
\end{align*}
Set $\eta_k = \eta$, $\gamma_k = \gamma = \frac{36 L^2 \eta }{ \mu^2 }$, $\beta = (18\mu \eta)^{1/2}$.
Recall that $\| \nabla P(w_k) - \nabla \phi_w^k \|^2 \leq \frac{2 L^2}{\mu} ( P(w_k) - \phi(w_k, a_k) )$.
By Lemma \ref{lemma:primal_gap_recursion} and \ref{lemma:auxiliary_recursion}, we have
\begin{align}\label{eq:combine_two_inequalities}
L_{k+1}
\leq &
P(w_k) - P^*
- \frac{\eta}{2} \| \nabla P(w_k) \|^2 + \frac{\eta}{2} \| \nabla P(w_k) - \nabla \phi_w^k \|^2 
\nonumber\\
&+ \frac{L_P \eta^2 \sigma^2}{2}
+ \frac{1}{8} \Bigg( 
(1 - \frac{\mu\gamma}{2}) \E[ P(w_k) - \phi(w_k, a_k) ]
\nonumber\\
&
+ (1 - \frac{\mu\gamma}{2}) ( \frac{\eta}{2} + 2\eta + L \eta^2 ) \| \nabla P(w_k) - \nabla \phi_w^k \|^2 
\nonumber\\
&
+ (1 - \frac{\mu\gamma}{2}) ( - \frac{\eta}{2} + 2\eta + L \eta^2) \| \nabla P(w_k) \|^2
\nonumber\\
&
+ L \sigma^2 ( \eta^2 + \gamma^2 (1 + 2 C_g^2) )
\Bigg)
\nonumber\\
&
+ \underbrace{ \Big( 2 \gamma C_h^2 L_g^2 + \Big( \frac{L^4 \eta }{\mu^5} \Big)^{1/2} 2 C_h^2 L_g^2 \Big) \E[ \| \omega_{k+1} - h(a_k) \|^2 ] }_{ = A }
\end{align}

Recall that $\eta \leq \frac{1}{4L}$. Before bounding term $A$, we first simplify the coefficients of $\E [ \| \nabla P(w_k) \|^2 ]$ and $\E [ \| \nabla P(w_k) - \nabla \phi_w(w_k, a_k) \|^2 ]$ as follows
\begin{align*}
&
\text{for } [ \| \nabla P(w_k) \|^2 ]:
\\
&
- \frac{\eta}{2} + \frac{1}{8} ( 1 - \frac{\mu \gamma}{2} ) ( -\frac{\eta}{2} + 2 \eta + L \eta^2 )
\leq 
- \frac{\eta}{4}  ,
\\
&
\text{for } \E [ \| \nabla P(w_k) - \nabla \phi_w(w_k, a_k) \|^2 ]:
\\
&
\frac{\eta}{2} + \frac{1}{8} ( 1 - \frac{\mu \gamma}{2} ) ( \frac{\eta}{2} + 2 \eta + L \eta^2 )
\leq 
\frac{7\eta}{8}  .
\end{align*}

Due to $L$-smoothness of $\phi$ in $a$ and Lemma \ref{lemma:PL_EB}, we can merge $\E [ \| \nabla P(w_k) - \nabla \phi_w(w_k, a_k) \|^2 ]$ with $P(w_k) - \phi(w_k, a_k)$.
Then the coefficient of $P(w_k) - \phi(w_k, a_k)$ in (\ref{eq:combine_two_inequalities}) can be derived as follows
\begin{align*}
\frac{1}{8} ( 1 - \frac{\mu \gamma}{2} ) + \frac{2L^2}{\mu} \cdot \frac{7\eta}{8}
=
\frac{1}{8} ( 1 - \frac{18 L^2 \eta}{\mu} + 14 \frac{L^2 \eta}{\mu} )
\leq 
\frac{1}{8} (1 - \frac{\mu \eta}{4})  ,
\end{align*}
where the inequality is due to $\mu \leq L$. Now we employ Lemma \ref{lemma:compositional_recursion} to bound term $A$ in (\ref{eq:combine_two_inequalities}) as follows
\begin{align}\label{eq:bounding_biased_term}
&
\Big( 2 \gamma_k C_h^2 L_g^2 + \Big( \frac{L^4 \eta }{\mu^5} \Big)^{1/2} 2 C_h^2 L_g^2 \Big) \E[ \| \omega_{k+1} - h(a_k) \|^2 ]
\nonumber\\
\leq &
2 C_h^2 L_g^2 L^2 / \mu^2 \sqrt{ \frac{\eta}{\mu} } ( 18 \sqrt{\mu \eta} + 1 ) ( 1-\beta ) \| \omega_k - h(a_k) \|^2
\nonumber\\
&
+ 2 C_h^2 L_g^2 L^2 / \mu^2 \sqrt{ \frac{\eta}{\mu} } ( 18 \sqrt{\mu \eta} + 1 ) ( \frac{C_h^2 \gamma^2 C_h^2 C_g^2 }{ \beta } + 2 \beta^2 \sigma^2 )
\nonumber\\
\leq &
2 C_h^2 L_g^2 L^2 / \mu^2 \sqrt{ \frac{\eta}{\mu} } ( 1 - \frac{\mu \eta}{4} ) \| \omega_k - h(a_k) \|^2
\nonumber\\
&
+ 20 C_h^2 L_g^2 L^2 / \mu^2 \sqrt{ \frac{\eta}{\mu} } ( \frac{ 36^2 C_h^2 C_g^2 L^4 \eta^2 }{18\mu^4 \sqrt{\mu\eta} } + 2 \sigma^2 18^2 \mu \eta )  .
\end{align}

By plugging the above simplified coefficients and merging (\ref{eq:bounding_biased_term}) into (\ref{eq:combine_two_inequalities}), we have the following
\begin{align*}
L_{k+1}
\leq &
P(w_k) - P^*
- \frac{\eta}{4} \| \nabla P(w_k) \|^2 
\\
&
+ \frac{1}{8} (1 - \frac{\mu \eta}{4}) ( P(w_k) - \phi(w_k, a_k) )
\\
&
+ \frac{L_P \eta^2 \sigma^2}{2}
+ \frac{1}{8} L \sigma^2 ( \eta^2 + \gamma^2 ( 1 + 2 C_g^2) )
\\
&
+ 2 C_h^2 L_g^2 L^2 / \mu^2 \sqrt{ \frac{\eta}{\mu} } ( 1 - \frac{\mu \eta}{4} ) \| \omega_k - h(a_k) \|^2
\\
&
+ 20 C_h^2 L_g^2 L^2 / \mu^2 \sqrt{ \frac{\eta}{\mu} } ( \frac{ 36^2 C_h^2 C_g^2 L^4 \eta^2 }{18\mu^4 \sqrt{\mu\eta} } + 2 \sigma^2 18^2 \mu \eta ) 
\\
\leq &
(1 - \frac{\mu\eta}{4}) ( P(w_k) - P^* )
+ \frac{1}{8} (1 - \frac{\mu \eta}{4}) ( P(w_k) - \phi(w_k, a_k) )
\\
&
+ 2 C_h^2 L_g^2 L^2 / \mu^2 \sqrt{ \frac{\eta}{\mu} } ( 1 - \frac{\mu \eta}{4} ) \| \omega_k - h(a_k) \|^2
+ B  
\\
= &
( 1 - \frac{\mu \eta}{4} ) L_k
+ B
\end{align*}
where the last inequality is due to $\mu$-PL condition of $P$ and we define
\begin{align*}
&
B 
:=
\frac{L_P \eta^2 \sigma^2}{2} + \frac{1}{8} L \sigma^2 ( \eta^2 + \gamma^2 ( 1 + 2 C_g^2) )
\\
&+ 20 C_h^2 L_g^2 L^2 / \mu^2 \sqrt{ \frac{\eta}{\mu} } ( \frac{ 36^2 C_h^2 C_g^2 L^4 \eta^2 }{18\mu^4 \sqrt{\mu\eta} } + 2 \sigma^2 18^2 \mu \eta )
\end{align*}

Finally, we expand the above recursion from $K$ to $1$ and have
\begin{align*}
L_K 
\leq
(1 - \frac{\mu \eta}{4})^K L_0
+ B \sum_{k=0}^{K-1} (1 - \frac{\mu \eta}{4})^k
\leq
\exp(-\frac{\mu \eta K}{4}) L_0 + \frac{4 B}{\mu \eta}
\end{align*}
We can verify that $\frac{4B}{\mu\eta} \leq 3\epsilon/4$ if $\eta$ satisfies the following three inequalities:
\begin{align*}
\eta \leq &
\frac{ \mu^4 \epsilon }{ 16 \sigma^2 (163 + 324 C_g^2) L^5 },
~~
\eta \leq
\frac{ \mu^8 }{ 23040 C_h^6 C_g^2 L_g^2 L^6 }
\\
\eta \leq &
\frac{ \mu^5 \epsilon^2 }{ ( 4 \cdot 160 \cdot 324 )^2 C_h^2 L_g^2 L^4 }
\end{align*}
In practice, $\epsilon$ is usually set to a very small value, so we employ the last inequality as the upper bound of $\eta$.
Then, to make $\exp(-\frac{\mu \eta K}{4})L_0 \leq \epsilon/4$, $K$ has the following lower bound.
\begin{align*}
K 
\geq
\frac{4}{\mu\eta} \log(4L_0 / \epsilon)
= 
\frac{ ( 8 \cdot 160 \cdot 324 )^2 C_h^2 L_g^2 L^4 }{ \mu^6 \epsilon^2 } \log(4L_0 / \epsilon) 
\end{align*}
Since $L_{k+1}$ is an upper bound of $\E[P(w_{k+1}) - P^*]$, to $\epsilon$-primal gap, we have the total number of iterations $K$ = $\tilde O\Big( \frac{ L^4 }{ \mu^6 \epsilon^2} \Big)$.

\vspace{1.0ex}

\subsection{ Proof of Theorem \ref{theorem:convergence_approximation_error} } 

To prove Theorem \ref{theorem:convergence_approximation_error}, let $\Delta_{K+1} = \E[\| \omega_{k+1} - h(a_k) \|^2]$ and use Lemma \ref{lemma:compositional_recursion}:
\begin{align*}
  \Delta_K
  \leq &
  (1-\beta) \Delta_{K-1}
  + \frac{ C_h^2 \gamma^2 C_h^2 C_g^2 }{\beta}
  + 2 \beta^2 \sigma^2
  \\
  \leq &
  (1-\beta)^{K-1} \Delta_1
  + \frac{ C_h^2 \gamma^2 C_h^2 C_g^2 }{\beta^2}
  + 2 \beta \sigma^2
  \leq \epsilon  ,
\end{align*}
where the second inequality is due to 
$\sum_{k=0}^{K-1} (1-\beta)^k \leq \frac{1}{\beta}$.
The last inequality is due to 
\begin{align*}
&
O( (1-\beta)^{K-1} )
\leq 
O(\exp(-\beta (K-1)))
\\
&
= 
O(\exp( - \sqrt{\mu \eta} \frac{L^6 \log(1/\epsilon) }{\mu^6 \epsilon^2} ))
\leq 
O( \epsilon^{1/\epsilon} )
\leq 
O(\epsilon),
\\
&
O(\frac{\gamma^2}{ \beta^2 } )
\leq
O(\frac{L^6 \eta^2}{\mu^4\mu\eta})
\leq 
O(\frac{L^6 \eta}{\mu^4\mu})
\leq
O(\epsilon) ,
\\
&
O(\beta)
\leq 
O(\sqrt{\mu \eta})
\leq 
O(\epsilon)  .
\end{align*}

\subsection{Proof of Lemma \ref{lemma:primal_gap_recursion} }

\begin{proof}
(of Lemma \ref{lemma:primal_gap_recursion})


Conditioned on iteration $k$, by $L_P$-smoothness of $P$, we have
\begin{align*}
&
\E [ P(w_{k+1}) - P(x_k) ]
\leq 
\E[ \langle \nabla P(w_k), w_{k+1} - w_k \rangle ]
+ \E [ \frac{L_P}{2} \| w_{k+1} - w_k \|^2 ]
\\
= &
\E[ \langle \nabla P(w_k), - \eta_k \nabla_w \phi_{i_1}(w_k, a_{i_1}) \rangle ]
+ \E [ \frac{L_P}{2} \| - \eta_k \nabla_w \phi_{i_1}(w_k, a_{i_1}) \|^2 ]
\\
\leq &
- \eta_k \langle \nabla P(w_k), \nabla_w \phi_{i_1}(w_k, a_{i_1}) \rangle
+ \frac{L_P \eta_k^2}{2} \E [ \| \nabla_w \phi(w_k, a) \|^2 ]
+ \frac{L_P \eta_k^2 \sigma^2}{2}
\\
\leq &
- \frac{\eta_k}{2} \| \nabla P(w_k) \|^2
+ \frac{\eta_k}{2} \| \nabla P(w_k) - \nabla_w \phi(w_k, a) \|^2
+ \frac{L_P \eta_k^2 \sigma^2}{2}  ,
\end{align*}
where the above first inequality is due to 
\begin{align*}
&
\E [ \| \nabla_w \phi_{i_1}(w_k, a_{i_1}) - \nabla_w \phi(w_k, a) + \nabla_w \phi(w_k, a) \|^2 ]
\\
= &
\E [ \| \nabla_w \phi_{i_1}(w_k, a_{i_1}) - \nabla_w \phi(w_k, a) \|^2 ]
+ \E [ \| \nabla_w \phi(w_k, a) \|^2 ]
\\
&
+ 2 \E [ \langle \nabla_w \phi_{i_1}(w_k, a_{i_1}) - \nabla_w \phi(w_k, a), \nabla_w \phi(w_k, a) \rangle ]
\\
\leq &
\sigma^2
+ \E [ \| \nabla_w \phi(w_k, a) \|^2 ]
\end{align*} 
and the second inequality is due to $\eta_k \leq 1/(4L_P)$.
\end{proof}

\subsection{Proof of Lemma \ref{lemma:auxiliary_recursion} }

\begin{proof}
(of Lemma \ref{lemma:auxiliary_recursion})

The following lemma is the standard analysis for one step update using stochastic gradient descent except that we keep the biased estimation term.
\begin{lemma}
\label{lemma:one_step_biased}
Suppose Assumption \ref{assumptions} hold.
For iteration $k$ of Algorithm \ref{algorithm:SCAGDA}, we have
\begin{align*}
&
P(w_{k+1}) - \phi(w_{k+1}, a_{k+1})
\\
\leq &
(1 - \frac{\mu\gamma_k}{2}) \E[ P(w_{k+1}) - \phi(w_{k+1}, a_k) ]
+ L \gamma_k^2 \sigma^2 (1 + 2 C_g^2)
\\
&
+ \gamma_k C_h^2 L_g^2 ( \frac{1}{2} (1-\frac{L\gamma_k}{2}) + 2 L\gamma_k ) \E[ \| \omega_{k+1} - h(a_k) \|^2 ].
\end{align*}
\end{lemma}

Now we can prove Lemma \ref{lemma:auxiliary_recursion}.
Denote $\nabla \phi_w^k = \nabla_w \phi(w_k, a_k)$ and $\hat \nabla \phi_w^k = \nabla_w \phi_{i_1}(w_k, a_{i_1}^k)$.

By Lemma \ref{lemma:one_step_biased}, we have
\begin{align*}
&
P(w_{k+1}) - \phi(w_{k+1}, a_{k+1})
\\
\leq &
(1 - \frac{\mu\gamma_k}{2}) \E[ P(w_{k+1}) - \phi(w_{k+1}, a_k) ]
\\
&
+ \gamma_k C_h^2 L_g^2 ( \frac{1}{2} (1-\frac{L\gamma_k}{2}) + 2 L\gamma_k ) \E[ \| \omega_{k+1} - h(a_k) \|^2 ]
+ L \gamma_k^2 \sigma^2 (1 + 2 C_g^2)
\\
\leq &
(1 - \frac{\mu\gamma_k}{2}) \E[ P(w_k) - \phi(w_k, a_k) + P(w_{k+1}) -  P(w_k) + \phi(w_k, a_k) 
\\
&
- \phi(w_{k+1}, a_k) ]
+ 2 \gamma_k C_h^2 L_g^2 \E[ \| \omega_{k+1} - h(a_k) \|^2 ]
+ L \gamma_k^2 \sigma^2 (1 + 2 C_g^2)
\\
\leq &
(1 - \frac{\mu\gamma_k}{2}) \E[ P(w_k) - \phi(w_k, a_k) ]
\\
&
+ (1 - \frac{\mu\gamma_k}{2}) ( - \frac{\eta_k}{2} \| \nabla P(w_k) \|^2 + \frac{\eta_k}{2} \| \nabla P(w_k) - \nabla \phi_w^k \|^2 + \frac{L \eta_k^2 \sigma^2 }{2} ) 
\\
&
+ (1 - \frac{\mu\gamma_k}{2}) \E [ ( \langle - \nabla \phi_w^k , w_{k+1} - w_k \rangle + \frac{L}{2} \| w_{k+1} - w_k \|^2 ) ]
\\
&
+ 2 \gamma_k C_h^2 L_g^2 \E[ \| \omega_{k+1} - h(a_k) \|^2 ]
+ L \gamma_k^2 \sigma^2 (1 + 2 C_g^2)
\\
= &
(1 - \frac{\mu\gamma_k}{2}) \E[ P(w_k) - \phi(w_k, a_k) ]
\\
&
+ (1 - \frac{\mu\gamma_k}{2}) ( - \frac{\eta_k}{2} \| \nabla P(w_k) \|^2 + \frac{\eta_k}{2} \| \nabla P(w_k) - \nabla \phi_w^k \|^2 + \frac{L \eta_k^2 \sigma^2 }{2} ) 
\\
&
+ (1 - \frac{\mu\gamma_k}{2}) ( \E [ \eta_k \langle \nabla \phi_w^k , \hat \nabla \phi_w^k \rangle ] + \frac{L \eta_k^2 }{2} \| \hat \nabla \phi_w^k -  \nabla \phi_w^k + \nabla \phi_w^k\|^2 )
\\
&
+ 2 \gamma_k C_h^2 L_g^2 \E[ \| \omega_{k+1} - h(a_k) \|^2 ]
+ L \gamma_k^2 \sigma^2 (1 + 2 C_g^2)
\\
&
+ \frac{L \eta_k^2 }{2} ( \| \hat \nabla \phi_w^k -  \nabla \phi_w^k \|^2 
+ 2 \| \nabla \phi_w^k - \nabla P(w_k) \|^2 + 2 \| \nabla P(w_k) \|^2 ) )
\end{align*}
\begin{align*}
&
+ 2 \gamma_k C_h^2 L_g^2 \E[ \| \omega_{k+1} - h(a_k) \|^2 ]
+ L \gamma_k^2 \sigma^2 (1 + 2 C_g^2)
\\
\leq &
(1 - \frac{\mu\gamma_k}{2}) \E[ P(w_k) - \phi(w_k, a_k) ]
\\
&
+ (1 - \frac{\mu\gamma_k}{2}) ( \frac{\eta_k}{2} + 2\eta_k + L \eta_k^2 ) \| \nabla P(w_k) - \nabla \phi_w^k \|^2 
\\
&
+ (1 - \frac{\mu\gamma_k}{2}) ( - \frac{\eta_k}{2} + 2\eta_k + L \eta_k^2) \nabla P(w_k) \|^2
\\
&
+ 2 \gamma_k C_h^2 L_g^2 \E[ \| \omega_{k+1} - h(a_k) \|^2 ]
+ L \sigma^2 ( \eta_k^2 + \gamma^2 (1 + 2 C_g^2) )
\end{align*}
where the third inequality is due to Lemma \ref{lemma:primal_gap_recursion} and $L$-smoothness of $\phi(w,a)$ in $a$.
The fourth inequality is due to $(a+b)^2 \leq 2a^2 + 2b^2$.
\end{proof}

\subsection{Proof of Lemma \ref{lemma:one_step_biased}}

\begin{proof}
(of Lemma \ref{lemma:one_step_biased})

Denote $\nabla \phi_a^k = \nabla_a \phi(w_{k+1}, a_k) = \nabla_a f(w_{k+1}, a_k) - \nabla h(a_k)^\top \nabla g(\omega_{k+1})$ 
and 
$\hat \nabla \phi_a^k = \nabla_a f_{i_2}(w_{k+1}, a_{i_2}^k) - \nabla h_{j_2}(a_{i_2}^k)^\top \nabla g(\omega_{i_2}^{k+1})$ for simplicity.
Note that $\E[\nabla_a f_{i_2}(w_{k+1}, a_{i_2}^k)] = \nabla_a f(w_{k+1}, a_k)$, $\E[\nabla h_{j_2}(a_{i_2}^k)] = \nabla h(a_k)$.
Then we have

\begin{align*}
&
\phi(w_{k+1}, a_k) - P(w_{k+1}) + P(w_{k+1}) - \phi(w_{k+1}, a_{k+1})
\\
= &
- \phi(w_{k+1}, a_{k+1}) - (- \phi(w_{k+1}, a_k) )
\\
\leq &
\langle -\nabla \phi_a^k , a_{k+1} - a_k \rangle + \frac{L}{2} \| a_{k+1} - a_k \|^2
\end{align*}
\begin{align*}
= &
\langle -\nabla \phi_a^k , \gamma_k \hat \nabla \phi_a^k \rangle + \frac{L \gamma_k^2}{2} \| \hat \nabla \phi_a^k \|^2
\\
= &
- \gamma_k \langle \nabla \phi_a^k , \nabla \phi_a^k - \nabla \phi_a^k + \gamma_k \hat \nabla \phi_a^k \rangle + \frac{L \gamma_k^2}{2} \| \nabla \phi_a^k - \nabla \phi_a^k +  \hat \nabla \phi_a^k \|^2
\\
= &
- \gamma_k \| \nabla \phi_a^k \|^2
- \gamma \langle \nabla \phi_a^k, \hat \nabla \phi_a^k - \nabla \phi_a^k \rangle
\\
&
+ \frac{L \gamma_k^2}{2} ( \| \hat \nabla \phi_a^k -  \nabla \phi_a^k \|^2 + \| \nabla \phi_a^k \|^2 + 2 \langle \hat \nabla \phi_a^k - \nabla \phi_a^k, \nabla \phi_a^k \rangle ) ,
\end{align*}

where the first inequality is due to $L$-smoothness of $\phi(w,a)$ in $a$.

Taking expectation over randomness of $i_2$ and $j_2$, we have
\begin{align*}
&
\E [ \phi(w_{k+1}, a_k) - P(w_{k+1}) + P(w_{k+1}) - \phi(w_{k+1}, a_{k+1})  ]
\\
\leq &
- \gamma_k (1 - \frac{L \gamma_k}{2}) \E [ \| \nabla \phi_a^k \|^2 ]
\\
- \gamma & (1 - \frac{L \gamma_k}{2} ) \langle \nabla \phi_a^k, \E [ \nabla h_{i_2, j_2}(a_{i_2}^k) ]^\top \E [ \nabla g(\omega_{i_2}^{k+1})] - {\nabla h(a_k)}^\top \nabla g(h(a_k)) \rangle
\\
&
+ L \gamma_k^2 \E \| \nabla_a f_{i_2}(w_{k+1}, a_{i_2}) - \nabla_a f(w_{k+1}, a_k) \|^2
\\
&
+ L \gamma_k^2 \E \| \nabla h_{j_2}(a_{i_2}^k)^\top \nabla g(\omega^{k+1}_{i_2}) - {\nabla h(a_k)}^\top \nabla g(h(a_k)) \|^2
\\
\leq &
- \gamma_k ( 1 - \frac{L\gamma_k}{2} - \frac{1}{2} (1 - \frac{L\gamma_k}{2}) ) \E[ \| \nabla \phi_a^k \|^2]
\\
&
+ \frac{\gamma_k}{2} (1-\frac{L\gamma_k}{2}) C_h^2 L_g^2 \E[ \| \omega_{k+1} - h(a_k) \|^2 ]
+ L \gamma_k^2 \sigma^2
\\
&
+ 2 L \gamma_k^2 ( C_g^2 \sigma^2 + C_h^2 L_g^2 \| \omega_{k+1} - h(a_k) \|^2 )  ,
\end{align*}

where the first inequality is due to $(a + b)^2 \leq 2a^2 + 2b^2$,the last inequality is due to Young's inequality $ab \leq a^2/2 + b^2/2$, $E[ \| \nabla h_{i_2, j_2}(a_{i_2}^k) - \nabla h(a_k) \|^2 ] \leq \sigma^2$ and $\E[ \| \nabla_a f_{i_2}(w_{k+1}, a_{i_2})\\ - \nabla_a f(w_{k+1}, a_k) \|^2 ] \leq \sigma^2$.

Finally, we finish the proof by setting $\gamma_k \leq 1 / (4 L)$ and the $\mu$-PL condition $2\mu ( P(w_{k+1}) - \phi(w_{k+1}, a_k) ) \leq \| \nabla \phi_a^k \|^2$.
\end{proof}

\subsection{Proof of Proposition \ref{prop:GAK_converge_to_DTW} }
\label{appendix:section:GAK_to_DTW}

\begin{proof}
(of Proposition \ref{prop:GAK_converge_to_DTW})

Suppose $\{ y_i \}_{i=1}^n$ where $y_i \geq 0$.
We have
\begin{align*}
\max_{i \in \{ 1,...,n \}} y_i
\leq
\sum_{i=1}^n y_i
\leq
n \max_{i \in \{ 1,...,n \}} y_i .
\end{align*}

Let $y_i = \exp( x_i / \nu )$ ($x_i$ can be negative), we have
\begin{align*}
\max_{i \in \{ 1,...,n \}} \exp( x_i / \nu )
\leq
\sum_{i=1}^n \exp( x_i / \nu )
\leq
n \max_{i \in \{ 1,...,n \}} \exp( x_i / \nu ) .
\end{align*}

Taking logarithm, we have
\begin{align*}
\log\Big( \max_{i \in \{ 1,...,n \}} \exp( x_i / \nu ) \Big)
\leq
\log\Big( \sum_{i=1}^n \exp( x_i / \nu ) \Big)
\leq &
\log\Big( n \max_{i \in \{ 1,...,n \}} \exp( x_i / \nu ) \Big)
\\
= &
\log(n) + \log\Big( \max_{i \in \{ 1,...,n \}} \exp( x_i / \nu ) \Big)  .
\end{align*}

Since $\exp(x)$ is a monotone increasing function, we have $\exp(x_1) < \exp(x_2)$ for $x_1 < x_2$. 
Applying it in the above inqualities, we have
$\log\Big( \max_{i \in \{ 1,...,n \}} \exp( x_i / \nu ) \Big) = \log\Big( \exp(  \max_{i \in \{ 1,...,n \}} x_i / \nu ) \Big)$, which leads to
\begin{align*}
\log\Big( \exp( \max_{i \in \{ 1,...,n \}} x_i / \nu ) \Big)
\leq
\log\Big( \sum_{i=1}^n \exp( x_i / \nu ) \Big)
\leq
\log(n) + \log\Big( \exp( \max_{i \in \{ 1,...,n \}} x_i / \nu ) \Big)  .
\end{align*}

By cancelling $\log$ with $\exp$ and multiplying $\nu$ in both sides, we have
\begin{align*}
\max_{i \in \{ 1,...,n \}} x_i 
\leq
\nu \log\Big( \sum_{i=1}^n \exp( x_i / \nu ) \Big)
\leq
\nu \log(n) + \max_{i \in \{ 1,...,n \}} x_i  .
\end{align*}

By replacing $x_i$ with $- d_{\pi} (x, x')$, we have
\begin{align*}
\max_{i \in \{ 1,...,n \}} - d_{\pi} (x, x')
\leq
\nu \log\Big( \sum_{i=1}^n \exp( - d_{\pi} (x, x') / \nu ) \Big)
\leq
\nu \log(n) + \max_{i \in \{ 1,...,n \}} - d_{\pi} (x, x')  .
\end{align*}

To better fit $D_{\GAK}$ and $D_{\DTW}$, we may also have the following equivalent inequalities:
\begin{align}\label{eq:GAK_to_DTW}
&
- \max_{i \in \{ 1,...,n \}} - d_{\pi} (x, x')
\geq
- \nu \log\Big( \sum_{i=1}^n \exp( - d_{\pi} (x, x') / \nu ) \Big)
\geq
- \nu \log(n) - \max_{i \in \{ 1,...,n \}} - d_{\pi} (x, x')  
\nonumber\\
\Rightarrow &
\nonumber\\
&
\underbrace{ \min_{i \in \{ 1,...,n \}} d_{\pi} (x, x') }_{ = D_{\DTW} }
\geq
\underbrace{ - \nu \log\Big( \sum_{i=1}^n \exp( - d_{\pi} (x, x') / \nu ) \Big) }_{ = D_{\GAK} }
\geq
- \nu \log(n) + \underbrace{ \min_{i \in \{ 1,...,n \}} d_{\pi} (x, x') }_{ = D_{\DTW} } .
\end{align}

\end{proof}

The above Equation (\ref{eq:GAK_to_DTW}) indicates the relationship between $D_{\GAK}$ and $D_{\DTW}$, as well as how fast $D_{\GAK}$ converges to $D_{\DTW}$ when $\nu$ decreases and approaches to $0$.

\vspace{1.0ex}

First, for a pair of time-series signals $(x, x')$, $D_{\DTW}$ (hard minimum) is always less than $D_{\GAK}$ (soft minimum).
Second, the gap between $D_{\DTW}$ and $D_{\GAK}$ can be upper bounded by $\nu \log(n)$, where $n = |\mathcal A|$ in our case, i.e., the total number of possible alignments between $x$ and $x'$.
Since the logarithm function $\log(n)$ is dominated by $\nu$, we may set $\nu$ to a very small value to approximate $D_{\DTW}$ using $D_{\GAK}$ by making $\nu \log(n)$ very small.
On the other hand, we can also tune $\nu$ in $D_{\GAK}$ to provide more flexibility, which sometimes leads to better practical performance.

\subsection{Stochastic Approximation in Optimization Problem of Equation (\ref{eq:min_max_general_form})  }
\label{appendix:section:stochastic_approximation}

Denote the dual variable $a_i$ as a whole: $a = [ a_1,\cdots, a_n ] \in \mathbb R^{C \times T \times n}$.
Then we can concatenate all components $h_{i, j}(a_i)$ for $i=1,\cdots,n$ and $j=1,\cdots,m$ together: 
\begin{align*}
h(a) = 
\frac{1}{n} \cdot
\left[
\begin{array}{cccc}
  \frac{1}{m} h_{1,1}(a_1), & \frac{1}{m} h_{1,2}(a_1), & ..., & \frac{1}{m} h_{1,m}(a_1)\\
  ...    \\
  \frac{1}{m} h_{i,1}(a_i), & \frac{1}{m} h_{i,2}(a_i), & ..., & \frac{1}{m} h_{i,m}(a_i)\\
  ...\\
  \frac{1}{m} h_{n,1}(a_n), & \frac{1}{m} h_{n,2}(a_n), & ..., & \frac{1}{m} h_{n,m}(a_n)
\end{array}
\right] 
\end{align*}
Note that if we sample an index $(i,j)$ to get $h(i,j)(a_i)$, then it is a stochastic approximation of $h(a)$. We can explicitly define the following notation:
\begin{align*}
\hat h_{i,j}(a_i) = 
\left[
\begin{array}{ccc}
  0, &  ..., & 0\\
  ... & ... & ...    \\
  0, & ..., h_{i,j}(a_i),..., & 0\\
  ... & ... & ...    \\
  0, &  ..., & 0
\end{array}
\right] 
\end{align*}
Then we can compute the expectation of $\hat h_{i,j}(a_i)$ as follows by sampling each $(i, j)$ index with $\frac{1}{mn}$ probability:
\begin{align*}
\sum_{i=1}^n \sum_{j=1}^m \frac{1}{m n} \hat h_{i, j}(a_i) = h(a)  ,
\end{align*}
This means that $\hat h_{i,j}(a_i)$ is an unbiased estimation of $h(a)$.
Similarly, we can find that $\hat h_{i,j}(a_i)$ is an unbiased estimation of $h_i(a_i)$. Same development can be done for $\nabla h(a)$.
\end{document}